\documentclass{article}

\usepackage{microtype}
\usepackage{graphicx}
\usepackage{subfigure}
\usepackage{booktabs} 

\usepackage{hyperref}

\usepackage[accepted]{icml2024}

\usepackage{amsmath}
\usepackage{amssymb}
\usepackage{mathtools}
\usepackage{amsthm}

\usepackage[capitalize,noabbrev]{cleveref}

\usepackage[utf8]{inputenc} 
\usepackage[T1]{fontenc}    
\usepackage{url}            
\usepackage{amsfonts}       
\usepackage{nicefrac}       
\usepackage{xcolor}         
\usepackage{color, colortbl}
\definecolor{greyC}{RGB}{180,180,180}
\definecolor{greyL}{RGB}{235,235,235}
\definecolor{shadecolor}{rgb}{0.92,0.92,0.92}
\definecolor{color1}{RGB}{255,190,122}
\definecolor{color2}{RGB}{142,207,201}
\definecolor{color3}{RGB}{190,184,220}
\definecolor{color4}{RGB}{130,176,210}
\usepackage{multicol}
\usepackage{multirow}
\usepackage{makecell}
\usepackage{wrapfig}
\usepackage{soul}
\usepackage{threeparttable}
\usepackage{dsfont}
\usepackage{enumerate}
\usepackage{amsmath,amsthm,amssymb}
\usepackage{natbib}
\usepackage{authblk}
\usepackage{wrapfig}
\usepackage{framed}
\usepackage{ulem}
\usepackage{rotating}
\usepackage{bbm}
\usepackage{pythonhighlight}
\theoremstyle{plain}

\theoremstyle{definition}

\theoremstyle{remark}

\usepackage[textsize=tiny]{todonotes}

\icmltitlerunning{MVMoE: Multi-Task Vehicle Routing Solver with Mixture-of-Experts}

\begin{document}

\twocolumn[
\icmltitle{MVMoE: \uline{M}ulti-Task \uline{V}ehicle Routing Solver with \uline{M}ixture-\uline{o}f-\uline{E}xperts}



\icmlsetsymbol{equal}{*}

\begin{icmlauthorlist}
\icmlauthor{Jianan Zhou}{ntu}
\icmlauthor{Zhiguang Cao}{smu}
\icmlauthor{Yaoxin Wu}{tue}
\icmlauthor{Wen Song}{shandong}
\icmlauthor{Yining Ma}{ntu}
\icmlauthor{Jie Zhang}{ntu}
\icmlauthor{Chi Xu}{ntu,simtech}
\end{icmlauthorlist}

\icmlaffiliation{ntu}{College of Computing and Data Science, Nanyang Technological University, Singapore}
\icmlaffiliation{tue}{Department of Information Systems, Eindhoven University of Technology, The Netherlands}
\icmlaffiliation{shandong}{Institute of Marine Science and Technology, Shandong University, China}
\icmlaffiliation{smu}{School of Computing and Information Systems, Singapore Management University, Singapore}
\icmlaffiliation{simtech}{Singapore Institute of Manufacturing Technology (SIMTech), Agency for Science, Technology and Research (A*STAR), Singapore}

\icmlcorrespondingauthor{Yaoxin Wu}{y.wu2@tue.nl}

\icmlkeywords{Machine Learning, ICML}

\vskip 0.3in
]



\printAffiliationsAndNotice{}  


\begin{abstract}
Learning to solve vehicle routing problems (VRPs) has garnered much attention. However, most neural solvers are only structured and trained independently on a specific problem, making them less generic and practical. In this paper, we aim to develop a unified neural solver that can cope with a range of VRP variants simultaneously. Specifically, we propose a multi-task vehicle routing solver with mixture-of-experts (MVMoE), which greatly enhances the model capacity without a proportional increase in computation. We further develop a hierarchical gating mechanism for the MVMoE, delivering a good trade-off between empirical performance and computational complexity. Experimentally, our method significantly promotes zero-shot generalization performance on 10 unseen VRP variants, and showcases decent results on the few-shot setting and real-world benchmark instances. We further conduct extensive studies on the effect of MoE configurations in solving VRPs, and observe the superiority of hierarchical gating when facing out-of-distribution data. The source code is available at: \url{https://github.com/RoyalSkye/Routing-MVMoE}.
\end{abstract}

\section{Introduction}
\label{intro}
Vehicle routing problems (VRPs) are a class of canonical combinatorial optimization problems (COPs) in operation research and computer science, with a wide spectrum of applications in logistics~\citep{cattaruzza2017vehicle}, transportation~\citep{wu2023neural}, and manufacturing~\citep{zhang2023review}. The intrinsic NP-hard nature makes VRPs exponentially expensive to be solved by exact solvers. As an alternative, heuristic solvers deliver suboptimal solutions within reasonable time, but need substantial domain expertise to be designed for each problem. Recently, learning to solve VRPs has received much attention~\citep{bengio2021machine,bogyrbayeva2024machine}, with fruitful neural solvers being developed. Most of them apply deep neural networks to learn solution construction policies via various training paradigms (e.g., reinforcement learning (RL)). Besides gaining decent performance, they are characterized by less computational overhead and domain expertise than conventional solvers. However, prevailing neural solvers still need network structures tailored and trained independently for each specific VRP, instigating prohibitive training overhead and less practicality when facing multiple VRPs. 

In this paper, we aim to develop a unified neural solver, which can be trained for solving a range of VRP variants simultaneously, and has decent zero-shot generalization capability on unseen VRPs. 
A few recent works explore similar problem settings. \citet{wang2023efficient} applies multi-armed bandits to solve multiple VRPs, while \citet{lin2024cross} adapts the model pretrained on one base VRP to target VRPs by efficient fine-tuning.
They fail to achieve zero-shot generalization to unseen VRPs due to the dependence on networks structured for predetermined problem variants.
\citet{anonymous2024multitask} empowers the neural solver with such generalizability by the compositional zero-shot learning~\citep{ruis2021independent}, which treats VRP variants as different combinations of a set of underlying attributes and uses a shared network to learn their representations.
However, it still leverages existing network structure proposed for simple VRPs, which is limited by its model capacity and empirical performance.

Motivated by the recent advance of large language models (LLMs)~\citep{kaplan2020scaling,floridi2020gpt,touvron2023llama}, we propose a multi-task VRP solver with mixture-of-experts (MVMoE). Typically, a mixture-of-
expert (MoE) layer replaces a feed-forward network (FFN) with several "experts" in a Transformer-based model, which are a group of FFNs with respective trainable parameters. 
An input to the MoE layer is routed to specific expert(s) by a gating network, and only parameters in selected expert(s) are activated (i.e., conditional computation~\citep{jacobs1991adaptive,jordan1994hierarchical}). In this manner, partially activated parameters can effectively enhance the model capacity without a proportional increase in computation, 
making the training and deployment of LLMs viable. 
Therefore, towards a more generic and powerful neural solver, we first propose an MoE-based neural VRP solver, and present a hierarchical gating mechanism for a good trade-off between empirical performance and computational complexity.
We choose the setting from \citet{anonymous2024multitask} as a test bed due to its potential to solve an exponential number of new VRP variants as any combination of the underlying attributes.

Our contributions are summarized as follows. 
1) We propose a unified neural solver MVMoE to solve multiple VRPs, which first brings MoEs into the study of COPs. The sole MVMoE can be trained on diverse VRP variants, and facilitate a strong zero-shot generalization capability on unseen VRPs.
2) We develop a hierarchical gating mechanism for MVMoE to attain a favorable balance between empirical performance and computational overhead. 
Surprisingly, it exhibits much stronger out-of-distribution generalization capability than the base gating.
3) Extensive experiments demonstrate that MVMoE significantly improves the zero-shot generalization against baselines
on 10 unseen VRP variants,
and achieves decent results on the few-shot setting and real-world instances.
We further provide extensive studies on the effect of MoE configurations (such as the position of MoEs, the number of experts, and the gating mechanism) on the zero-shot generalization performance.

\section{Related Work}
\label{related_work}
\textbf{Neural VRP Solvers.}
Two mainstreams exist in literature on learning to solve VRPs: 
1) \emph{Construction-based solvers}, which learn policies to construct solutions in an end-to-end manner. \citet{vinyals2015pointer} proposes Pointer Network to estimate the optimal solution to the traveling salesman problem (TSP) in an autoregressive way. The follow-up works apply RL to explore better approximate solutions to TSP~\cite{bello2017neural} and capacitated vehicle routing problem (CVRP)~\citep{nazari2018reinforcement}. \citet{kool2018attention} proposes an attention-based model (AM) that uses Transformer to solve a series of VRPs independently. By leveraging the symmetry property in solutions, \citet{kwon2020pomo} proposes the policy optimization with multiple optima (POMO) to further promote the performance in solving TSP and CVRP. Other construction-based solvers are often developed on top of AM and POMO~\citep{kwon2021matrix,li2021deep,kim2022symnco,berto2023rl4co,chen2023neural,grinsztajn2023winner,chalumeau2023combinatorial,hottung2024polynet}. Besides the autoregressive manner, several works construct a heatmap to solve VRPs in a non-autoregressive manner~\citep{joshi2019efficient,fu2021generalize,kool2022deep,qiu2022dimes,sun2023difusco,min2023unsupervised,ye2023deepaco,kim2024ant}.
2) \emph{Improvement-based solvers,} which learn policies to iteratively refine an initial solution until a termination condition is satisfied. The policies are often trained in contexts of classic local search~\citep{croes1958method,shaw1998using} or specialized heuristic solvers~\citep{helsgaun2017extension} for obtaining more efficient or effective search components ~\citep{chen2019learning,lu2020learning,hottung2020neural,d2020learning,wu2021learning,xin2021neurolkh,hudson2022graph,zhou2023learning,ma2023learning}.
In general, construction-based solvers can efficiently achieve desired performance, whereas improvement-based solvers have the potential to deliver better solutions given prolonged inference time.

Recent research uncovers the deficient generalization capability of neural solvers, which suffer from drastic performance decrement on unseen data~\citep{joshi2021learning}. Previous works mainly focus on the cross-size generalization~\citep{fu2021generalize,hou2023generalize,son2023meta,luo2023neural,drakulic2023bq} or cross-distribution generalization~\citep{zhang2022learning,geisler2022generalization,bi2022learning,jiang2023ensemblebased} or both~\citep{manchanda2022generalization,zhou2023towards,wang2024asp} on a single problem. In this paper, we step further to explore the generalization across different VRP variants~\citep{wang2023efficient,anonymous2024multitask,lin2024cross}.

\textbf{Mixture-of-Experts.}
The original idea of MoEs was proposed three decades ago~\citep{jacobs1991adaptive,jordan1994hierarchical}. In early concepts, the expert was defined as an entire neural network, and hence MoEs was similar to an ensemble of neural networks. 
\citet{eigen2013learning} launchs the era when researchers start applying MoEs as components of neural networks. 
As an early success of MoEs applied in large neural networks, \citet{shazeer2017} introduces the sparsely-gated MoEs in language modeling and machine translation, achieving state-of-the-art results at the time with only minor losses in computational efficiency. Follow-up works mainly focus on improving the gating mechanism~\citep{lewis2021base,roller2021hash,zuo2022taming,zhou2022mixture,puigcerver2023sparse,xue2024openmoe} or applications to other domains~\citep{lepikhin2020gshard,riquelme2021scaling,fedus2022switch}. We refer interested readers to \citet{yuksel2012twenty,fedus2022review} for a comprehensive survey.

\section{Preliminaries}
\label{prelim}
In this section, we first present the definition of CVRP, and then introduce its variants featured by additional constraints. 
Afterwards, we delineate recent construction-based neural solvers for VRPs~\citep{kool2018attention,kwon2020pomo}.

\textbf{VRP Variants.} We define a CVRP instance of size $n$ over a graph $\mathcal{G}=\{\mathcal{V}, \mathcal{E}\}$, where $\mathcal{V}$ includes a depot node $v_0$ and customer nodes $\{v_i\}_{i=1}^{n}$, and $\mathcal{E}$ includes edges $e(v_i, v_j)$ between node $v_i$ and $v_j (i \neq j)$. Each customer node is associated with a demand $\delta_i$, and a capacity limit $Q$ is set for each vehicle.
The solution (i.e., tour) $\tau$ is represented as a sequence of nodes, consisting of multiple sub-tours. Each sub-tour represents that a vehicle starts from the depot, visits a subset of customer nodes and returns to the depot. The solution is feasible if each customer node is visited exactly once, and the total demand in each sub-tour does not exceed the capacity limit $Q$.
We consider the Euclidean space with the cost function $c(\cdot)$ defined as the total length of the tour. The objective is to find the optimal tour $\tau^*$ with the minimal cost: $\tau^* = \arg\min_{\tau\in \Phi} c(\tau|\mathcal{G})$, where $\Phi$ is the discrete search space that contains all feasible tours.

\begin{figure}[!t]
    \vskip 0.1in
    \begin{center}
    \centerline{\includegraphics[width=0.99\columnwidth]{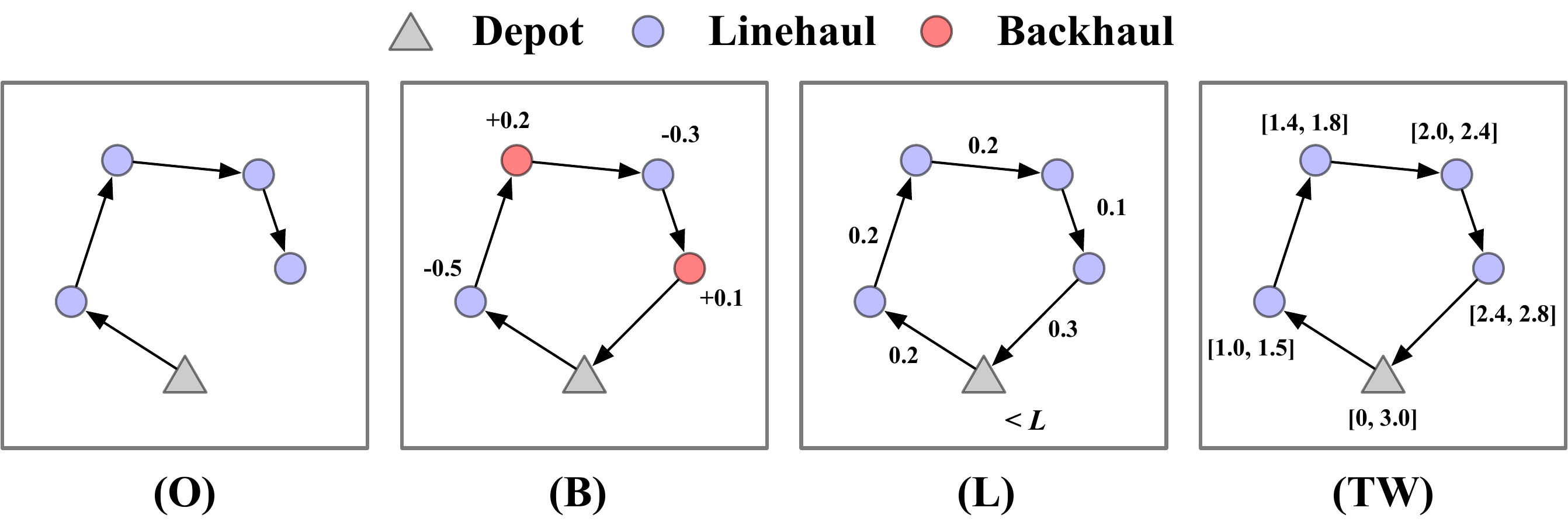}}
    \caption{Illustrations of sub-tours with various constraints: open route (O), backhaul (B), duration limit (L), and time window (TW).}
    \label{vrps}
    \end{center}
    \vskip -0.3in
\end{figure}

On top of CVRP (featured by the capacity constraint \emph{(C)}), several VRP variants involve additional practical constraints. 
1) \emph{Open Route (O):} The vehicle does not need to return to the depot $v_0$ after visiting customers; 
2) \emph{Backhaul (B):} The demand $\delta_i$ is positive in CVRP, representing a vehicle unloads goods at the customer node. In practice, a customer can have a negative demand, requiring a vehicle to load goods. We name the customer nodes with $\delta_i > 0$ as linehauls and the ones with $\delta_i < 0$ as backhauls. Hence, VRP with backhaul allows the vehicle traverses linehauls and backhauls in a mixed manner, without strict precedence between them;
3) \emph{Duration Limit (L):} To maintain a reasonable workload, the cost (i.e., length) of each route is upper bounded by a predefined threshold;
4) \emph{Time Window (TW):} Each node $v_i \in \mathcal{V}$ is associated with a time window $[e_i, l_i]$ and a service time $s_i$. A vehicle must start serving customer $v_i$ in the time slot from $e_i$ to $l_i$. If the  vehicle arrives earlier than $e_i$, it has to wait until $e_i$. All vehicles must return to the depot $v_0$ no later than $l_0$. The aforementioned constraints are illustrated in Fig.~\ref{vrps}.
By combining them, we can obtain 16 typical VRP variants, which are summarized in Table \ref{problem}.
Note that the combination is not a trivial addition of different constraints. For example, when the open route is coupled with the time window, the vehicle does not need to return to the depot, and hence the constraint imposed by $l_0$ at the depot is relaxed.
We present more details of VRP variants and the associated data generation process in Appendix~\ref{app:vrps}.

\textbf{Learning to Solve VRPs.} Typical neural solvers~\citep{kool2018attention, kwon2020pomo} parameterize the solution construction policy  by an attention-based neural network $\pi_{\theta}$, which is trained to generate a solution in an autoregressive way. The feasibility of the generated solution is guaranteed by the masking mechanism during decoding. Without loss of generality, we consider RL training paradigm, wherein the solution construction process is formulated as a Markov Decision Process (MDP). 
Given an input instance, the encoder processes it and attains all node embeddings, which, with the context representation of the constructed partial tour, represent the current state. The decoder takes them as inputs and outputs the probabilities of valid nodes (i.e., actions) to be selected. After a complete solution $\tau$ is constructed, its probability can be factorized via the chain rule such that $p_{\theta}(\tau|\mathcal{G}) = \prod_{t=1}^{T} p_{\theta} (\pi_{\theta}^{(t)}|\pi_{\theta}^{(<t)},\mathcal{G})$, where $\pi_{\theta}^{(t)}$ and $\pi_{\theta}^{(<t)}$ denote the selected node and constructed partial tour at step $t$, and $T$ is the number of total steps. The reward is defined as the negative tour length, i.e., $\mathcal{R} = -c(\tau|\mathcal{G})$.
Given a baseline function $b(\cdot)$ for training stability, the policy network $\pi_\theta$ is often trained by REINFORCE~\citep{williams1992simple} algorithm, which applies estimated gradients of the expected reward to optimize the policy as below,
\begin{equation}
    \label{eq:reinforce}
    \small
    \nabla_{\theta} \mathcal{L}_a(\theta|\mathcal{G}) = \mathbb{E}_{p_{\theta}(\tau|\mathcal{G})} [(c(\tau)-b(\mathcal{G})) \nabla_{\theta}\log p_{\theta}(\tau|\mathcal{G})].
\end{equation}

\section{Methodology}
\label{methodology}
\begin{figure*}[!t]
    \begin{center}
    \centerline{\includegraphics[width=1.98\columnwidth]{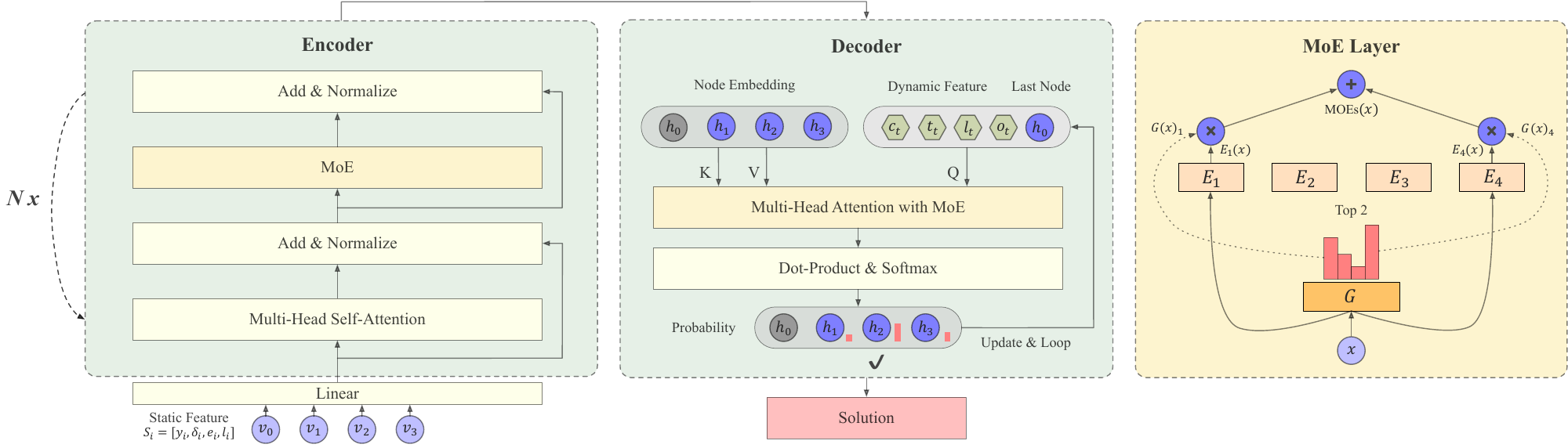}}
    \vskip -0.05in
    \caption{The model structure of MVMoE. [\emph{Green part}]: Given an input instance, the encoder and decoder output node embeddings and probabilities of nodes to be selected, respectively.
    The gray nodes are masked to satisfy problem-specific constraints for feasibility. The node with a deeper color denote a later node embedding.
    [\emph{Yellow part}]: In an MoE layer, where we take the (node-level) input-choice Top2 gating as an example, 
    the input $x$ (i.e., node)
    is routed to two experts that derive the two largest probabilities from the gating network $G$.}
    \label{moe}
    \end{center}
    \vskip -0.3in
\end{figure*}

In this section, we present the multi-task VRP solver with MoEs (MVMoE), and introduce the gating mechanism. Without loss of generality, we aim to learn a construction-based neural solver~\citep{kool2018attention,kwon2020pomo} for tackling VRP variants with the five constraints introduced in Section~\ref{prelim}. The structure of MVMoE is illustrated in Fig.~\ref{moe}. 


\subsection{Multi-Task VRP Solver with MoEs}

\textbf{Multi-Task VRP Solver.} Given an instance of a specific VRP variant, the \emph{static} features of each node $v_i$ are expressed by $\mathcal{S}_i=\{y_i, \delta_i, e_i, l_i\}$, where $y_i, \delta_i, e_i, l_i$ denote the coordinate, demand, start and end time of the time window, respectively. The encoder takes these static node features as inputs, and outputs $d$-dimensional node embeddings $h_i$. At the $t_{\rm{th}}$ decoding step, the decoder takes as input the node embeddings and a context representation, including the embedding of the last selected node and \emph{dynamic} features $\mathcal{D}_t=\{c_t, t_t, l_t, o_t\}$, where $c_t, t_t, l_t, o_t$ denote the remaining capacity of the vehicle, the current time, the length of the current partial route, and the presence indicator of the open route, respectively. Thereafter, the decoder outputs the probability distribution of nodes, from which a valid node is selected and appended to the partial solution. A complete solution is constructed in an autoregressive manner by iterating the decoding process.

In each training step, we randomly select a VRP variant, and train the neural network to solve associated instances in a batch. In this way, MVMoE is able to learn a unified policy that can tackle different VRP tasks. If only a subset of static or dynamic features are involved in the current selected VRP variant, the other features are padded to the default values (e.g., zeros). For example, given a CVRP instance, the static features of the $i_{\rm{th}}$ customer node are $\mathcal{S}_i^{(C)}=\{y_i,\delta_i,0,0\}$, and the dynamic features at the $t_{\rm{th}}$ decoding step are $\mathcal{D}_t^{(C)}=\{c_t,0,l_t,0\}$. 
In summary, motivated by the fact that different VRP variants may include some common attributes (e.g., coordinate, demand), we define the static and dynamic features as the union set of attributes that exist in all VRP variants. By training on a few VRP variants with these attributes, the policy network has the potential to solve unseen variants, which are characterized by different combinations of these attributes, i.e., the zero-shot generalization capability~\citep{anonymous2024multitask}.




\textbf{Mixture-of-Experts.} Typically, an MoE layer consists of 1) $m$ experts $\{E_1, E_2, \dots, E_m\}$, each of which is a linear layer or FFN with independent trainable parameters, and 2) a gating network $G$ parameterized by $W_G$, which decides how the inputs are distributed to experts.
Given a single input $x$, $G(x)$ and $E_j(x)$ denote the output of the gating network (i.e., an $m$-dimensional vector), and the output of the $j_{\rm{th}}$ expert, respectively. In light of this, the output of an MoE layer is calculated as,
\begin{equation}
\begin{aligned}
    \label{eq:moe}
    \text{MoE}(x) &= \sum_{j=1}^m G(x)_j E_j(x).
\end{aligned}
\end{equation}
Intuitively, a sparse vector $G(x)$ only activates a small subset of experts with partial model parameters, and hence saves the computation. Typically, a Top$K$ operator 
can achieve such sparsity
by only keeping the $K$-largest values while setting others as the negative infinity. In this case, the gating network calculates the output as $G(x) = \text{Softmax}(\text{Top}K(x\cdot W_G))$.
Given the fact that larger sparse models do not always lead to better performance~\citep{zuo2022taming}, it is crucial yet tricky to \emph{design effective and efficient gating mechanisms to endow each expert being sufficiently trained, given enough training data.} To this effect, some works have been put forward in language and vision domains, such as designing an auxiliary loss~\citep{shazeer2017} or formulating it as a linear assignment problem~\citep{lewis2021base} in pursuit of the load balancing. 

\textbf{MVMoE.} By integrating the above parts, we obtain the multi-task VRP solver with MoEs. The overall model structure is shown in Fig.~\ref{moe}, where we employ MoEs in both the encoder and decoder. In specific, we substitute MoEs for the FFN layer in the encoder, and substitute MoEs for the final linear layer of multi-head attention in the decoder.
We refer more details of the structure of MVMoE to Appendix \ref{app:model_structure}.
We empirically find our design is effective in generating high-quality solutions, and especially employing MoEs in the decoder tends to exert a greater influence on performance (see Section \ref{exps_moe}).
We jointly optimize all trainable parameters $\Theta$,
with the objective formulated as follows,
\begin{equation}
    \label{eq:loss}
    \min_{\Theta} \mathcal{L} = \mathcal{L}_a + \alpha \mathcal{L}_b,
\end{equation}
where $\mathcal{L}_a$ denotes the original loss function of the VRP solver
(e.g., the REINFORCE loss used to train the policy for solving VRP variants in Eq.~(\ref{eq:reinforce})), $\mathcal{L}_b$ denotes the loss function associated with MoEs (e.g., the auxiliary loss used to ensure load balancing in Eq.~(\ref{eq:aux_loss}) in Appendix~\ref{app:model_structure}), and $\alpha$ is a hyperparameter to control its strength. 

\begin{figure}[!t]
    \vskip 0.05in
    \begin{center}
    \centerline{\includegraphics[width=0.95\columnwidth]{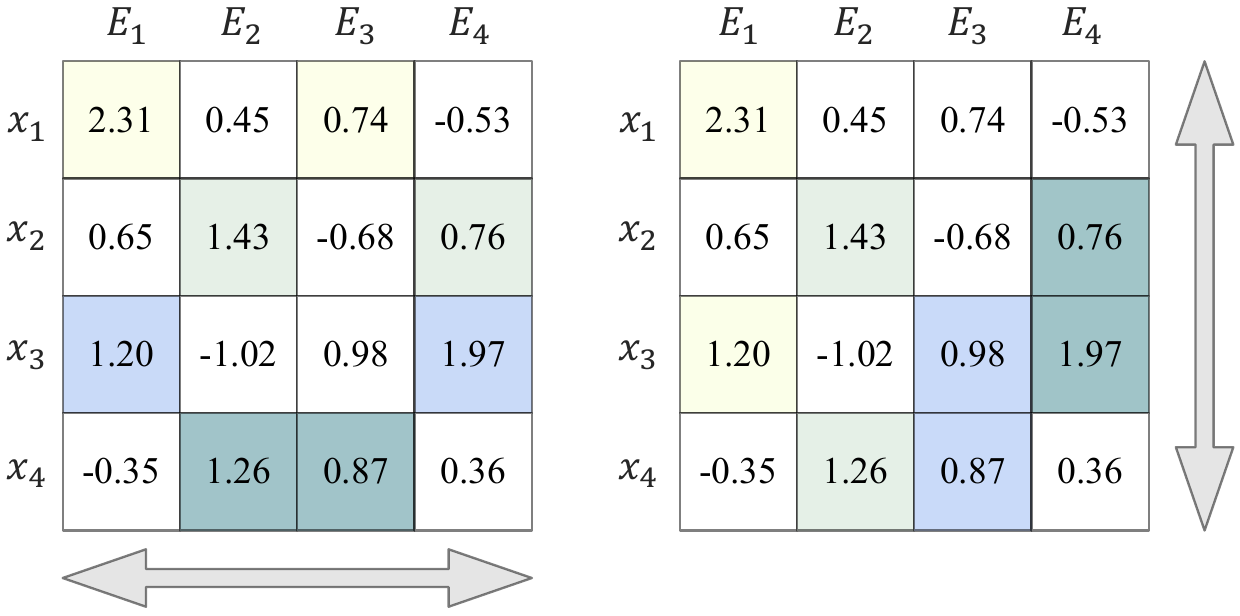}}
    \caption{An illustration of the score matrix and gating algorithm.
    \emph{Left panel:} Input-choice gating. \emph{Right panel:} Expert-choice gating. The selected experts or nodes are in color. The arrow marks the dimension, along which the Top$K$ experts or nodes are selected.}
    \label{gating}
    \end{center}
    \vskip -0.3in  
\end{figure}

\subsection{Gating Mechanism}
We mainly consider the node-level (or token-level) gating, by which each node is routed independently to experts.\footnote{In addition, we also investigate another two gating levels, i.e., instance-level and problem-level gating, which are presented in Section \ref{exps_moe} and Appendix \ref{app:model_structure}.} 
In each MoE layer, the extra computation originates from the forward pass of the gating network and the distribution of nodes to the selected experts.
While employing MoEs in the decoder can significantly improve the performance, the number of decoding steps $T$ increases as the problem size $n$ scales up. It suggests that compared to the encoder with a fixed number of gating steps $N$ ($\ll T$), applying MoEs in the decoder may substantially increase the computational complexity. In light of this, we propose a hierarchical gating mechanism to make the better use of MoEs in the decoder for gaining a good trade-off between empirical performance and computational complexity. Next, we detail the node-level and hierarchical gating mechanism.

\textbf{Node-Level Gating.}
The node-level gating routes inputs at the granularity of nodes. Let $d$ denote the hidden dimension and $W_G\in \mathbb{R}^{d\times m}$ denote trainable parameters of the gating network in MVMoE. Given a batch of inputs $X \in \mathbb{R}^{I\times d}$, where $I$ is the total number of nodes (i.e., batch size $B$ $\times$ problem scale $n$), each node is routed to the selected experts based on the score matrix $H = (X \cdot W_G) \in \mathbb{R}^{I\times m}$ predicted by the gating network. We illustrate an example of the score matrix in Fig.~\ref{gating}, where $x_i$ denotes the $i_{\rm{th}}$ node, and $E_j$ denotes the $j_{\rm{th}}$ expert in the node-level gating.

In this paper, we mainly consider two popular gating algorithms~\citep{shazeer2017,zhou2022mixture}:
1) \emph{Input-choice gating:} Each node selects Top$K$ experts based on $H$. Typically, $K$ is set to 1 or 2 to retain a reasonable computational complexity.
The input-choice gating is illustrated in the left panel of Fig.~\ref{gating}, where each node is routed to two experts with the largest scores (i.e., Top$2$). However, this method cannot guarantee load balancing. An expert may receive much more nodes than the others, resulting in a dominant expert while leaving others underfitting. To address this issue, most works employ an auxiliary loss to equalize quantities of nodes sent to different experts during training. 
Here we use the importance \& load loss \cite{shazeer2017} as $\mathcal{L}_b$ in Eq. (\ref{eq:loss}) to mitigate load imbalance (see Appendix~\ref{app:model_structure}).
2) \emph{Expert-choice gating:} Each expert selects Top$K$ nodes based on $H$. Typically, $K$ is set to $\frac{I\times \beta}{m}$, where $\beta$ is the capacity factor reflecting the average number of experts utilized by a node. 
The expert-choice gating is illustrated in the right panel of Fig.~\ref{gating}, where each expert selects two nodes with the largest scores given $\beta=2$. While this gating algorithm explicitly ensures load balancing, some nodes may not be chosen by any expert. We refer more details of the above gating algorithms to Appendix~\ref{app:model_structure}.

\begin{figure}[!t]
    \vskip 0.05in
    \begin{center}
    \centerline{\includegraphics[width=0.90\columnwidth]{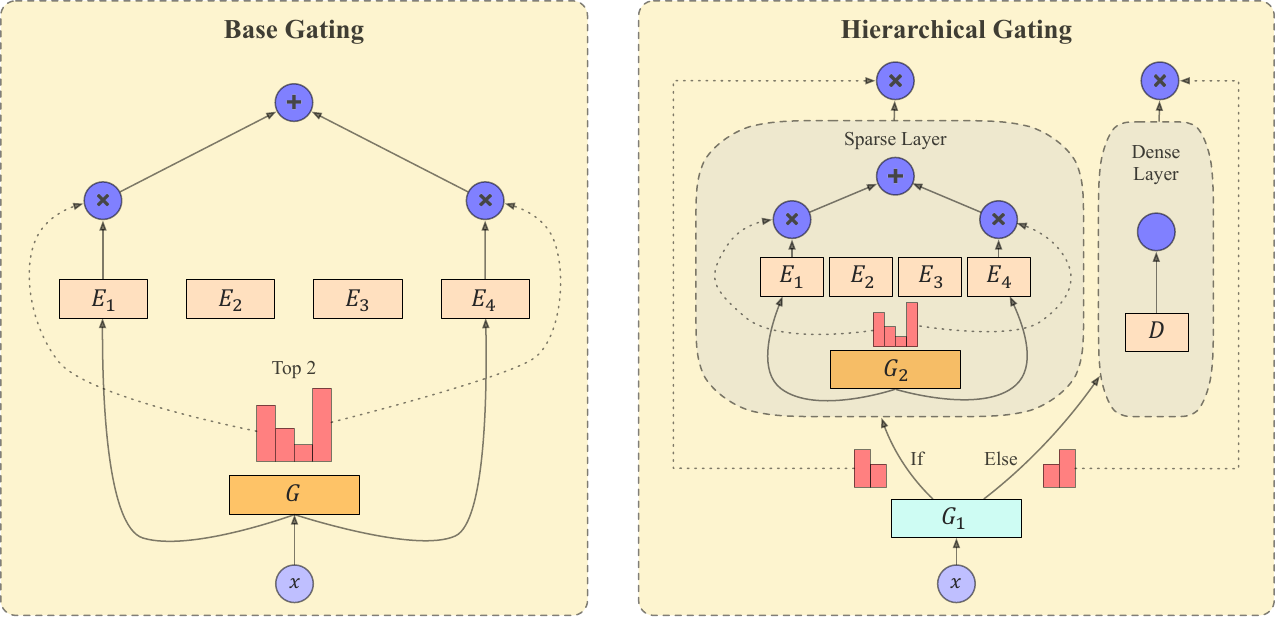}}
    \caption{A base gating (i.e., the input-choice gating with $K=2$) and its hierarchical gating counterpart. In the latter, the gating network $G_1$ routes inputs to the sparse layer ($\{G_2, E_1, E_2, E_3, E_4\}$) or the dense layer $D$. If the sparse layer is chosen, the gating network $G_2$ routes nodes to experts accoring to the base gating.
    }
    \label{hierarchical_gating}
    \end{center}
    \vskip -0.35in  
\end{figure}

\begin{table*}[!t]
  \vskip -0.05in
  \caption{Performance on 1K test instances of trained VRPs. * represents 0.000\%, with which the gaps are computed.}
  \label{exp_1}
  \vskip 0.1in
  \begin{center}
  \begin{small}
  \renewcommand\arraystretch{1.05}  
  \resizebox{0.98\textwidth}{!}{ 
  \begin{tabular}{ll|cccccc|ll|cccccc}
    \toprule
    \midrule
    \multicolumn{2}{c|}{\multirow{2}{*}{Method}} & \multicolumn{3}{c}{\textbf{$n=50$}} & \multicolumn{3}{c|}{$n=100$} & \multicolumn{2}{c|}{\multirow{2}{*}{Method}} &
    \multicolumn{3}{c}{\textbf{$n=50$}} & \multicolumn{3}{c}{$n=100$} \\
     & & Obj. & Gap & Time & Obj. & Gap & Time & & & Obj. & Gap & Time & Obj. & Gap & Time \\
    \midrule
    \multirow{8}*{\rotatebox{90}{CVRP}} & HGS & 10.334 & * & 4.6m & 15.504 & * & 9.1m & \multirow{8}*{\rotatebox{90}{VRPTW}} & HGS & 14.509 & * & 8.4m & 24.339 & * & 19.6m \\
    & LKH3 & 10.346 & 0.115\% & 9.9m & 15.590 & 0.556\% & 18.0m & & LKH3 & 14.607 & 0.664\% & 5.5m & 24.721 & 1.584\% & 7.8m \\
    & OR-Tools & 10.540 & 1.962\% & 10.4m & 16.381 & 5.652\% & 20.8m & & OR-Tools & 14.915 & 2.694\% & 10.4m & 25.894 & 6.297\% & 20.8m \\
    & OR-Tools (x10) & 10.418 & 0.788\% & 1.7h & 15.935 & 2.751\% & 3.5h & & OR-Tools (x10) & 14.665 & 1.011\% & 1.7h & 25.212 & 3.482\% & 3.5h \\
    & POMO & 10.418 & 0.806\% & 3s & 15.734 & 1.488\% & 9s & & POMO & 14.940 & 2.990\% & 3s & 25.367 & 4.307\% & 11s \\
    & POMO-MTL & 10.437 & 0.987\% & 3s & 15.790 & 1.846\% & 9s & & POMO-MTL & 15.032 & 3.637\% & 3s & 25.610 & 5.313\% & 11s \\
    & MVMoE/4E & \textbf{10.428} & \textbf{0.896\%} & 4s & \textbf{15.760} & \textbf{1.653\%} & 11s & & MVMoE/4E & \textbf{14.999} & \textbf{3.410\%} & 4s & \textbf{25.512} & \textbf{4.903\%} & 12s \\
    & MVMoE/4E-L & 10.434 & 0.955\% & 4s & 15.771 & 1.728\% & 10s & & MVMoE/4E-L & 15.013 & 3.500\% & 3s & 25.519 & 4.927\% & 11s \\
    \midrule
    \multirow{7}*{\rotatebox{90}{OVRP}} & LKH3 & 6.511 & 0.198\% & 4.5m & 9.828 & * & 5.3m & \multirow{7}*{\rotatebox{90}{VRPL}} & LKH3 & 10.571 & 0.790\% & 7.8m & 15.771 & * & 16.0m \\
    & OR-Tools & 6.531 & 0.495\% & 10.4m & 10.010 & 1.806\% & 20.8m & & OR-Tools & 10.677 & 1.746\% & 10.4m & 16.496 & 4.587\% & 20.8m \\
    & OR-Tools (x10) & 6.498 & * & 1.7h & 9.842 & 0.122\% & 3.5h & & OR-Tools (x10) & 10.495 & * & 1.7h & 16.004 & 1.444\% & 3.5h \\
    & POMO & 6.609 & 1.685\% & 2s & 10.044 & 2.192\% & 8s & & POMO & 10.491 & -0.008\% & 2s & 15.785 & 0.093\% & 9s \\
    & POMO-MTL & 6.671 & 2.634\% & 2s & 10.169 & 3.458\% & 8s & & POMO-MTL & 10.513 & 0.201\% & 2s & 15.846 & 0.479\% & 9s \\
    & MVMoE/4E & \textbf{6.655} & \textbf{2.402\%} & 3s & \textbf{10.138} & \textbf{3.136\%} & 10s & & MVMoE/4E & \textbf{10.501} & \textbf{0.092\%} & 3s & \textbf{15.812} & \textbf{0.261\%} & 10s \\
    & MVMoE/4E-L & 6.665 & 2.548\% & 3s & 10.145 & 3.214\% & 9s & & MVMoE/4E-L & 10.506 & 0.131\% & 3s & 15.821 & 0.323\% & 10s \\
    \midrule
    \multirow{6}*{\rotatebox{90}{VRPB}} & OR-Tools & 8.127 & 0.989\% & 10.4m & 12.185 & 2.594\% & 20.8m & \multirow{6}*{\rotatebox{90}{OVRPTW}} & OR-Tools & 8.737 & 0.592\% & 10.4m & 14.635 & 1.756\% & 20.8m \\
    & OR-Tools (x10) & 8.046 & * & 1.7h & 11.878 & * & 3.5h & & OR-Tools (x10) & 8.683 & * & 1.7h & 14.380 & * & 3.5h \\
    & POMO & 8.149 & 1.276\% & 2s & 11.993 & 0.995\% & 7s & & POMO & 8.891 & 2.377\% & 3s & 14.728 & 2.467\% & 10s \\
    & POMO-MTL & 8.182 & 1.684\% & 2s & 12.072 & 1.674\% & 7s & & POMO-MTL & 8.987 & 3.470\% & 3s & 15.008 & 4.411\% & 10s \\
    & MVMoE/4E & \textbf{8.170} & \textbf{1.540\%} & 3s & \textbf{12.027} & \textbf{1.285\%} & 9s & & MVMoE/4E & \textbf{8.964} & \textbf{3.210\%} & 4s & \textbf{14.927} & \textbf{3.852\%} & 11s \\
    & MVMoE/4E-L & 8.176 & 1.605\% & 3s & 12.036 & 1.368\% & 8s & & MVMoE/4E-L & 8.974 & 3.322\% & 4s & 14.940 & 3.941\% & 10s \\
    \midrule
    \bottomrule
  \end{tabular}}
  \end{small}
  \end{center}
  \vskip -0.1in
\end{table*}

\textbf{Hierarchical Gating.} In the VRP domain, it is computationally expensive to employ MoEs in each decoding step, since 1) the number of decoding steps $T$ increases as the problem size $n$ rises; 
2) the problem-specific feasibility constraints must be satisfied during decoding. To tackle the challenges, we propose to employ MoEs only in \emph{partial} decoding steps. Accordingly, we present a hierarchical gating, which learns to effectively and efficiently utilize MoEs during decoding. 

We illustrate the proposed hierarchical gating in Fig.~\ref{hierarchical_gating}. An MoE layer with the hierarchical gating includes two gating networks $\{G_1, G_2\}$, $m$ experts $\{E_1, E_2, \dots, E_m\}$, and a dense layer $D$ (e.g., a linear layer). Given a batch of inputs $X \in \mathbb{R}^{I\times d}$, the hierarchical gating routes them in two stages. In the first stage, $G_1$ decides to distribute inputs $X$ to either the sparse or dense layer according to the problem-level representation $X_1$. In specific, we obtain $X_1$ by applying the mean pooling along the first dimension of $X$, and process it to obtain the score matrix $H_1=(X_1\cdot W_{G_1}) \in \mathbb{R}^{1\times 2}$. Then, we route the batch of inputs $X$ to the sparse or dense layer by sampling from the probability distribution $G_1(X) = \text{Softmax}(H_1)$. 
Here we employ the problem-level gating in $G_1$ for the generality and efficiency of the hierarchical gating (see Appendix \ref{app:discussion} for further discussions).
In the second stage, if $X$ is routed to the sparse layer, the gating network $G_2$ is activated to route nodes to experts on the node-level by using aforementioned gating algorithms (e.g., the input-choice gating). Otherwise, $X$ is routed to the dense layer $D$ and transformed into $D(X) \in \mathbb{R}^{I\times d}$.
In summary, the hierarchical gating learns to output $G_1(X)_0 \sum_{j=1}^m G_2(X)_j E_j(X)$ or $G_1(X)_1 D(X)$ based on both problem-level and node-level representations.

Overall, the hierarchical gating improves the computational efficiency with a minor loss on the empirical performance. To balance the efficiency and performance of MVMoE, we use the base gating in the encoder and its hierarchical gating counterpart in the decoder. Note that the hierarchical gating is applicable to different gating algorithms, such as the input-choice gating~\citep{shazeer2017} and expert-choice gating~\citep{zhou2022mixture}. We also explore a more advanced gating algorithm~\citep{puigcerver2023sparse} for reducing the number of routed nodes and thus the computational complexity. But its empirical performance is unsatisfactory in the VRP domain (see Section~\ref{exps_analyses}).


\section{Experiments}
\label{exps}
In this section, we empirically verify the superiority of the proposed MVMoE, and provide insights into the application of MoEs to solve VRPs. We consider 16 VRP variants with five constraints. Due to page limit, we present more experimental results in Appendix \ref{app:exps}. All experiments are conducted on a machine with NVIDIA Ampere A100-80GB GPU cards and AMD EPYC 7513 CPU at 2.6GHz. 

\begin{table*}[!t]
  \vskip -0.05in
  \caption{Zero-shot generalization on 1K test instances of unseen VRPs. * represents 0.000\%, with which the gaps are computed.}
  \label{exp_2}
  \vskip 0.1in
  \begin{center}
  \begin{small}
  \renewcommand\arraystretch{1.05}  
  \resizebox{0.98\textwidth}{!}{ 
  \begin{tabular}{ll|cccccc|ll|cccccc}
    \toprule
    \midrule
    \multicolumn{2}{c|}{\multirow{2}{*}{Method}} & \multicolumn{3}{c}{\textbf{$n=50$}} & \multicolumn{3}{c|}{$n=100$} & \multicolumn{2}{c|}{\multirow{2}{*}{Method}} &
    \multicolumn{3}{c}{\textbf{$n=50$}} & \multicolumn{3}{c}{$n=100$} \\
     & & Obj. & Gap & Time & Obj. & Gap & Time & & & Obj. & Gap & Time & Obj. & Gap & Time \\
        \midrule
    \multirow{5}*{\rotatebox{90}{OVRPB}} & OR-Tools & 5.764 & 0.332\% & 10.4m & 8.522 & 1.852\% & 20.8m & \multirow{5}*{\rotatebox{90}{OVRPL}} & OR-Tools & 6.522 & 0.480\% & 10.4m & 9.966 & 1.783\% & 20.8m \\
    & OR-Tools (x10) & 5.745 & * & 1.7h & 8.365 & * & 3.5h & & OR-Tools (x10) & 6.490 & * & 1.7h & 9.790 & * & 3.5h \\
    & POMO-MTL & 6.116 & 6.430\% & 2s & 8.979 & 7.335\% & 8s & & POMO-MTL & 6.668 & 2.734\% & 2s & 10.126 & 3.441\% & 9s \\
    & MVMoE/4E & \textbf{6.092} & \textbf{5.999\%} & 3s & \textbf{8.959} & \textbf{7.088\%} & 9s & & MVMoE/4E & \textbf{6.650} & \textbf{2.454\%} & 3s & \textbf{10.097} & \textbf{3.148\%} & 10s \\
    & MVMoE/4E-L & 6.122 & 6.522\% & 3s & 8.972 & 7.243\% & 9s & & MVMoE/4E-L & 6.659 & 2.597\% & 3s & 10.106 & 3.244\% & 9s \\
    \midrule
    \multirow{5}*{\rotatebox{90}{VRPBL}} & OR-Tools & 8.131 & 1.254\% & 10.4m & 12.095 & 2.586\% & 20.8m & \multirow{5}*{\rotatebox{90}{VRPBTW}} & OR-Tools & 15.053 & 1.857\% & 10.4m & 26.217 & 2.858\% & 20.8m \\
    & OR-Tools (x10) & 8.029 & * & 1.7h & 11.790 & * & 3.5h & & OR-Tools (x10) & 14.771 & * & 1.7h & 25.496 & * & 3.5h \\
    & POMO-MTL & 8.188 & 1.971\% & 2s & 11.998 & 1.793\% & 8s & & POMO-MTL & 16.055 & 8.841\% & 3s & 27.319 & 7.413\% & 10s \\
    & MVMoE/4E & \textbf{8.172} & \textbf{1.776\%} & 3s & \textbf{11.945} & \textbf{1.346\%} & 9s & & MVMoE/4E & \textbf{16.022} & \textbf{8.600\%} & 4s & \textbf{27.236} & \textbf{7.078\%} & 11s \\
    & MVMoE/4E-L & 8.180 & 1.872\% & 3s & 11.960 & 1.473\% & 9s & & MVMoE/4E-L & 16.041 & 8.745\% & 4s & 27.265 & 7.190\% & 10s \\
    \midrule
    \multirow{5}*{\rotatebox{90}{VRPLTW}} & OR-Tools & 14.815 & 1.432\% & 10.4m & 25.823 & 2.534\% & 20.8m & \multirow{5}*{\rotatebox{90}{OVRPBL}} & OR-Tools & 5.771 & 0.549\% & 10.4m & 8.555 & 2.459\% & 20.8m \\
    & OR-Tools (x10) & 14.598 & * & 1.7h & 25.195 & * & 3.5h & & OR-Tools (x10) & 5.739 & * & 1.7h & 8.348 & * & 3.5h \\
    & POMO-MTL & 14.961 & 2.586\% & 3s & 25.619 & 1.920\% & 12s & & POMO-MTL & 6.104 & 6.306\% & 2s & 8.961 & 7.343\% & 8s \\
    & MVMoE/4E & \textbf{14.937} & \textbf{2.421\%} & 4s & \textbf{25.514} & \textbf{1.471\%} & 13s & & MVMoE/4E & \textbf{6.076} & \textbf{5.843\%} & 3s & \textbf{8.942} & \textbf{7.115\%} & 9s \\
    & MVMoE/4E-L & 14.953 & 2.535\% & 4s & 25.529 & 1.545\% & 12s & & MVMoE/4E-L & 6.104 & 6.310\% & 3s & 8.957 & 7.300\% & 9s \\
    \midrule
    \multirow{5}*{\rotatebox{90}{OVRPBTW}} & OR-Tools & 8.758 & 0.927\% & 10.4m & 14.713 & 2.268\% & 20.8m & \multirow{5}*{\rotatebox{90}{OVRPLTW}} & OR-Tools & 8.728 & 0.656\% & 10.4m & 14.535 & 1.779\% & 20.8m \\
    & OR-Tools (x10) & 8.675 & * & 1.7h & 14.384 & * & 3.5h & & OR-Tools (x10) & 8.669 & * & 1.7h & 14.279 & * & 3.5h \\
    & POMO-MTL & 9.514 & 9.628\% & 3s & 15.879 & 10.453\% & 10s & & POMO-MTL & 8.987 & 3.633\% & 3s & 14.896 & 4.374\% & 11s \\
    & MVMoE/4E & \textbf{9.486} & \textbf{9.308\%} & 4s & \textbf{15.808} & \textbf{9.948\%} & 11s & & MVMoE/4E & \textbf{8.966} & \textbf{3.396\%} & 4s & \textbf{14.828} & \textbf{3.903\%} & 12s \\
    & MVMoE/4E-L & 9.515 & 9.630\% & 3s & 15.841 & 10.188\% & 10s & & MVMoE/4E-L & 8.974 & 3.488\% & 4s & 14.839 & 3.971\% & 10s \\
    \midrule
    \multirow{5}*{\rotatebox{90}{VRPBLTW}} & OR-Tools & 14.890 & 1.402\% & 10.4m & 25.979 & 2.518\% & 20.8m & \multirow{5}*{\rotatebox{90}{OVRPBLTW}} & OR-Tools & 8.729 & 0.624\% & 10.4m & 14.496 & 1.724\% & 20.8m \\
    & OR-Tools (x10) & 14.677 & * & 1.7h & 25.342 & * & 3.5h & & OR-Tools (x10) & 8.673 & * & 1.7h & 14.250 & * & 3.5h \\
    & POMO-MTL & 15.980 & 9.035\% & 3s & 27.247 & 7.746\% & 11s & & POMO-MTL & 9.532 & 9.851\% & 3s & 15.738 & 10.498\% & 10s \\
    & MVMoE/4E & \textbf{15.945} & \textbf{8.775\%} & 4s & \textbf{27.142} & \textbf{7.332\%} & 12s & & MVMoE/4E & \textbf{9.503} & \textbf{9.516\%} & 4s & \textbf{15.671} & \textbf{10.009\%} & 11s \\
    & MVMoE/4E-L & 15.963 & 8.915\% & 4s & 27.177 & 7.473\% & 11s & & MVMoE/4E-L & 9.518 & 9.682\% & 4s & 15.706 & 10.263\% & 10s \\
    \midrule
    \bottomrule
  \end{tabular}}
  \end{small}
  \end{center}
  \vskip -0.15in
\end{table*}

\textbf{Baselines.} \emph{Traditional solvers:} We employ HGS~\citep{vidal2022hybrid} to solve CVRP and VRPTW instances with default hyperparameters (i.e., the maximum number of iterations without improvement is 20000). We run LKH3~\citep{helsgaun2017extension} to solve CVRP, OVRP, VRPL and VRPTW instances with 10000 trails and 1 run. OR-Tools~\citep{ortools_routing} is an open source solver for complex optimization problems. It is more versatile than LKH and HGS, and can solve all 16 VRP variants considered in this paper. We use the parallel cheapest insertion as the first solution strategy, and use the guided local search as the local search strategy in OR-Tools.
For $n=50/100$, we set the search time limit as 20s/40s to solve an instance, and also provide its results given 200s/400s (i.e., OR-Tools (x10)). For all traditional solvers, we use them to solve 32 instances in parallel on 32 CPU cores following \citet{kool2018attention}.

\emph{Neural solvers:} We compare our method to POMO~\citep{kwon2020pomo} and POMO-MTL~\citep{anonymous2024multitask}. While POMO is trained on each single VRP, POMO-MTL is trained on multiple VRPs by multi-task learning. 
Note that POMO-MTL is the dense model counterpart of MVMoE, which is structured by dense layers (e.g., FFNs) rather than sparse MoEs. In specific, POMO-MTL and MVMoE/4E possess 1.25M and 3.68M parameters, but they activate a similar number of parameters for each single input.

\textbf{Training.} We follow most setups in \cite{kwon2020pomo}. 1) \emph{For all neural solvers:} Adam optimizer is used with the learning rate of $1e-4$, the weight decay of $1e-6$, and the batch size of 128. The model is trained for 5000 epochs, with each containing 20000 training instances (i.e., 100M training instances in total). The learning rate is decayed by 10 for the last 10\% training instances. We consider two problem scales $n\in \{50, 100\}$ during training, according to \cite{anonymous2024multitask}.
2) \emph{For multi-task solvers:} The training problem set includes CVRP, OVRP, VRPB, VRPL, VRPTW, and OVRPTW (see Appendix \ref{app:train_vrps} for further discussions). In each batch of training, we randomly sample a problem from the set and generate its instances. Please refer to Appendix~\ref{app:vrps} for details of the generation procedure.
3) \emph{For our method:} We employ $m=4$ experts with $K=\beta=2$ in each MoE layer, and set the the weight $\alpha$ of the auxiliary loss $\mathcal{L}_b$ as 0.01. The default gating mechanism of MVMOE/4E is the node-level input-choice gating in both the encoder and decoder layers. MVMoE/4E-L is a computationally light version that replaces the input-choice gating with its hierarchical gating counterpart in the decoder.

\textbf{Inference.} For all neural solvers, we use greedy rollout with x8 instance augmentation following \citet{kwon2020pomo}. We report the average results (i.e., objective values and gaps) over the test dataset that contains 1K instances, and the total time to solve the entire test dataset. The gaps are computed with respect to the results of the best-performing traditional VRP solvers (i.e., * in Tables \ref{exp_1} and \ref{exp_2}). 

\begin{figure}[!t]
    \centering
    \centerline{\includegraphics[width=0.45\columnwidth]{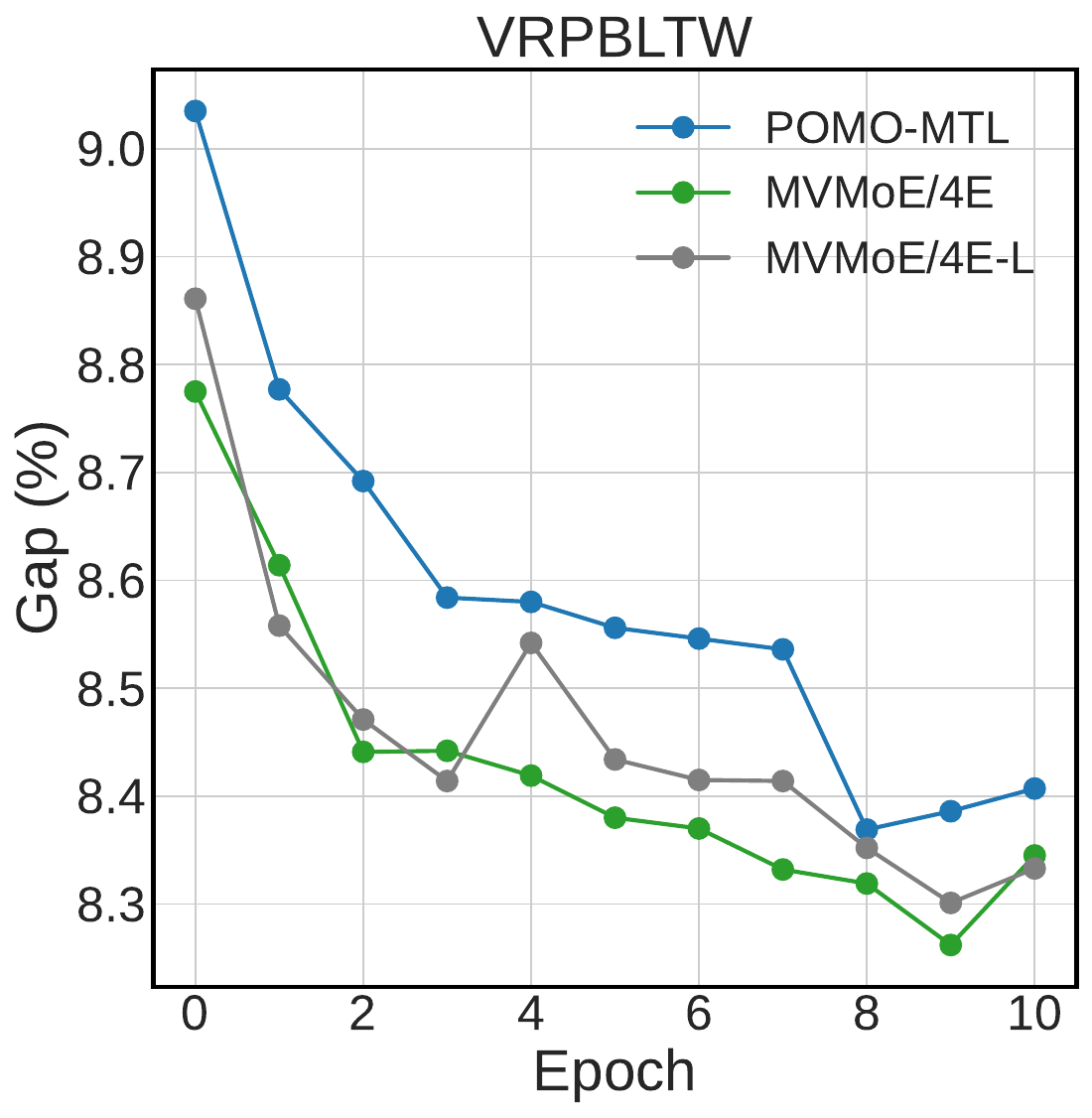}
    \hspace{1mm}
    \includegraphics[width=0.45\columnwidth]{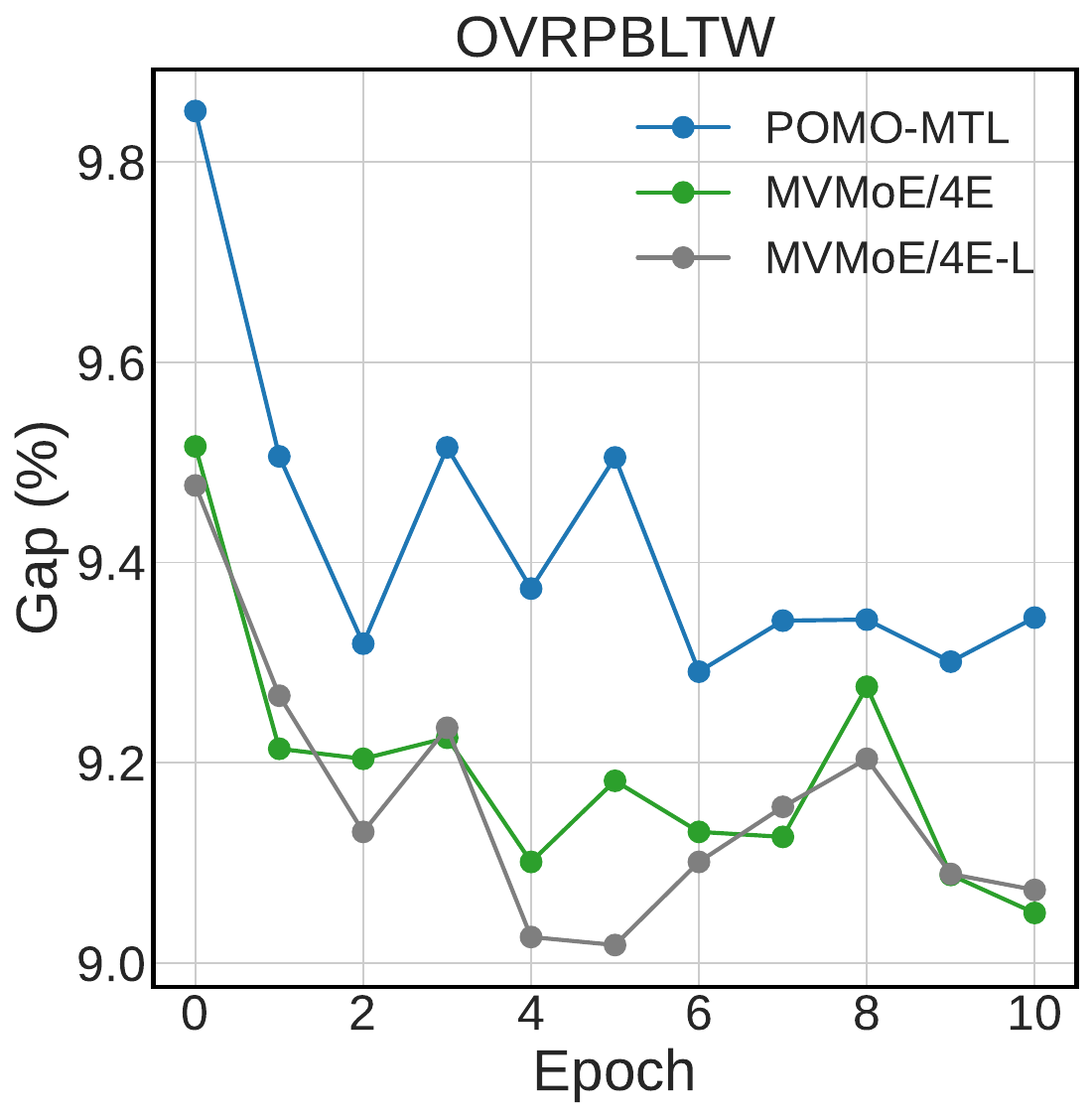}}
    \caption{Few-shot generalization on unseen VRPs.}
    \label{few_shot}
    \vskip -0.15in
\end{figure}

\begin{figure*}[!t]
    \vskip 0.05in
    \centering
    \centerline{
    \includegraphics[width=0.4\columnwidth]{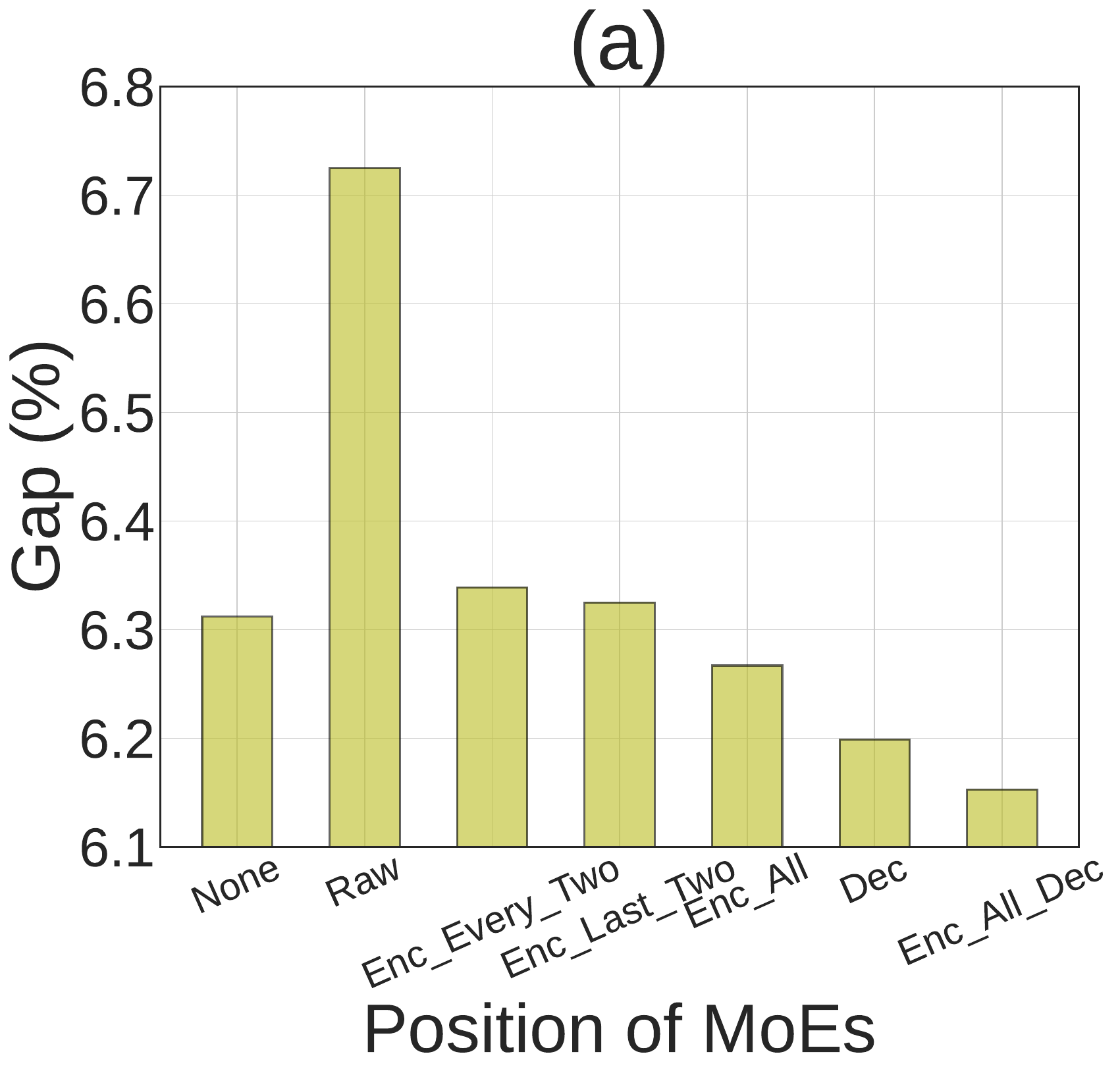}  
    \includegraphics[width=0.4\columnwidth]{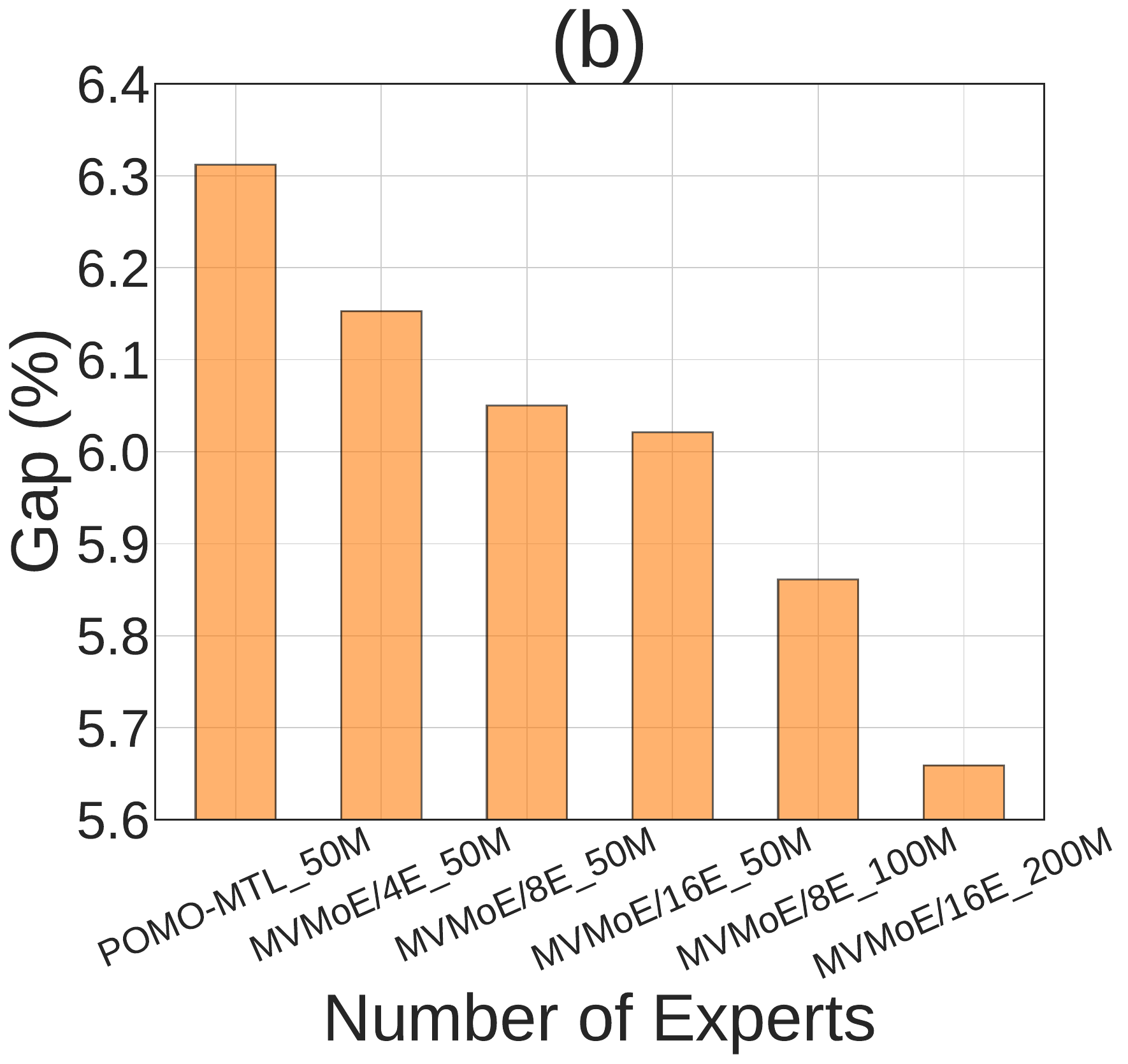}  
    \includegraphics[width=0.415\columnwidth]{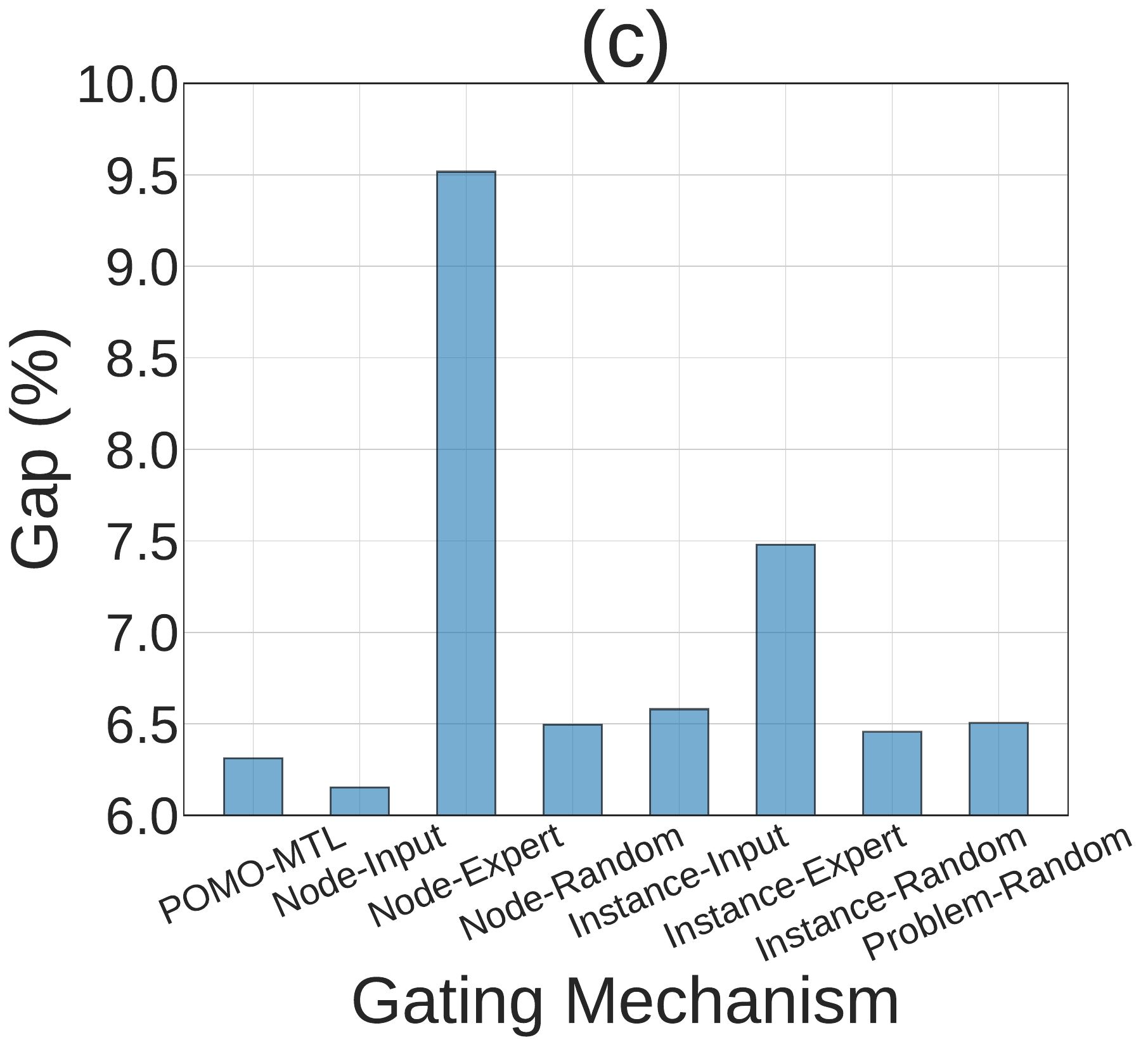}  
    \includegraphics[width=0.4\columnwidth]{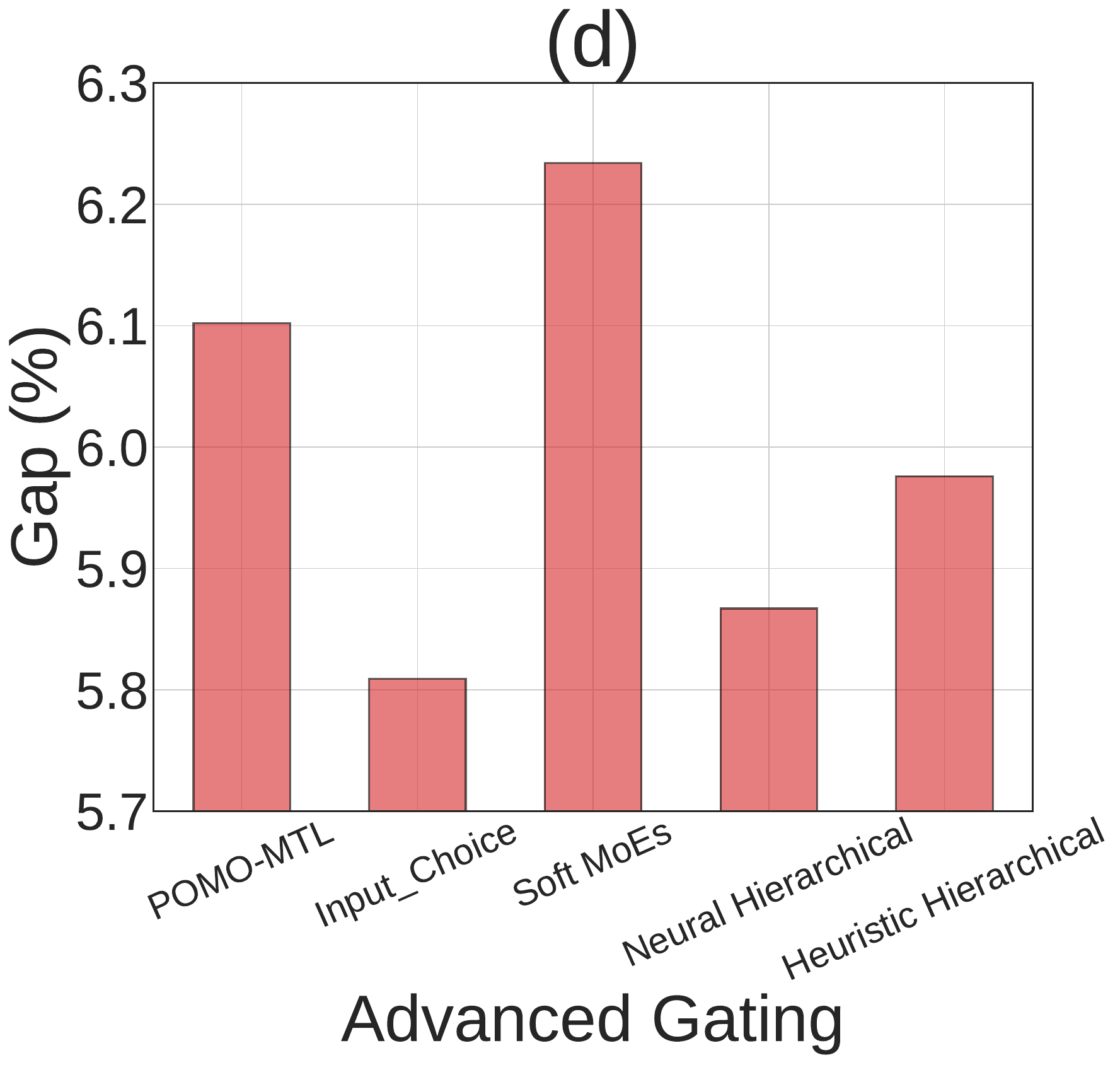}
    \includegraphics[width=0.4\columnwidth]{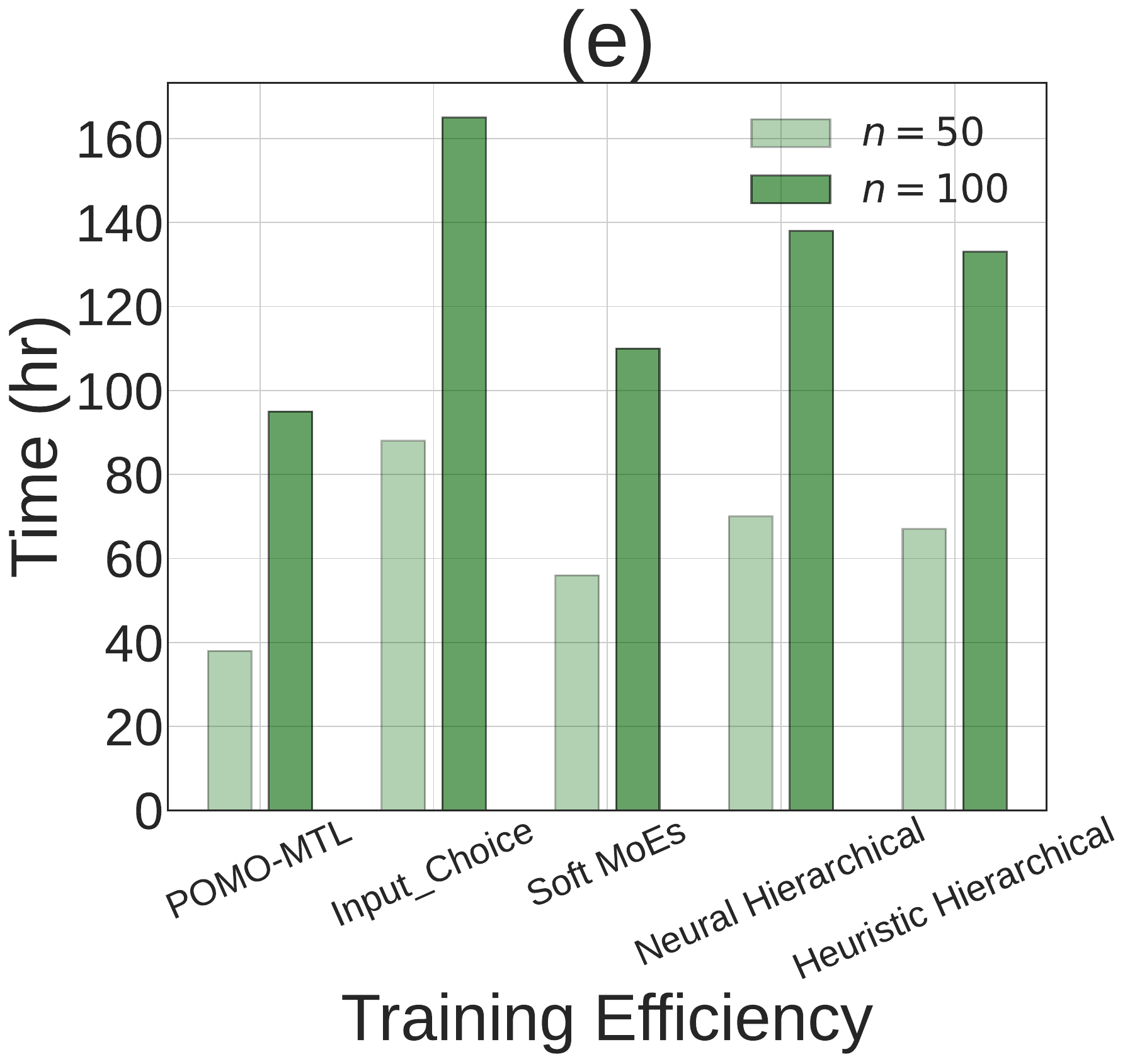}
    }
    \caption{\emph{Left three panels:} The effect of MoE settings on the average zero-shot generalization performance - \textbf{(a)} the position of MoEs; \textbf{(b)} the number of experts; \textbf{(c)} the gating mechanism.
    \emph{Right two panels:} Further analyses - \textbf{(d)} average zero-shot generalization performance of each method employing various gating algorithms in the decoder; \textbf{(e)} training efficiency of each gating algorithm.
    }
    \label{moe_setting}
\end{figure*}

\subsection{Empirical Results}
\label{exps_results}
\textbf{Performance on Trained VRPs.} We evaluate all methods on 6 trained VRPs and gather all results in Table~\ref{exp_1}. 
The single-task neural solver (i.e., POMO) achieves better performance than multi-task neural solvers on each single problem, since it is restructured and retrained on each VRP independently. However, its average performance over all trained VRPs is quite inferior as shown in Table~\ref{effect_constraints} in Appendix \ref{app:exps}, since each trained POMO is overfitted to a specific VRP. For example, the average performance of POMO solely trained on CVRP is 16.815\%, while POMO-MTL and MVMoE/4E achieve 2.102\% and 1.925\%, respectively. Notably, our neural solvers consistently outperform POMO-MTL. MVMoE/4E performs slightly better than MVMoE/4E-L at the expense of more computation. Despite that, MVMoE/4E-L exhibits stronger out-of-distribution generalization capability than MVMoE/4E (see Tables~\ref{exp_benchmark} and \ref{exp_benchmark_large_scale} in Appendix \ref{app:exps}).

\textbf{Generalization on Unseen VRPs.} We evaluate multi-task solvers on 10 unseen VRP variants. 
1) \emph{Zero-shot generalization:} We directly test the trained solvers on unseen VRPs. The results in Table~\ref{exp_2} reveal that the proposed MVMoE significantly outperforms POMO-MTL across all VRP variants.
2) \emph{Few-shot generalization:} We also consider the few-shot setting on $n=50$, where a trained solver is fine-tuned on the target VRP using 10K instances (0.01\% of total training instances) in each epoch. Without loss of generality, we conduct experiments on VRPBLTW and OVRPBLTW following the training setups. The results in Fig. \ref{few_shot} showcase MVMoE generalizes more favorably than POMO-MTL.

\subsection{Ablation on MoEs}
\label{exps_moe}
Here we explore the effect of different MoE settings on the zero-shot generalization of neural solvers, and provide insights on how to effectively apply MoEs to solve VRPs. Due to the fast convergence, we reduce the number of epochs to 2500 on VRPs of the size $n=50$, while leaving other setups unchanged. We set MVMoE/4E as the default baseline, and ablate on different components of MoEs below.

\textbf{Position of MoEs.} We consider three positions to apply MoEs in neural solvers: 1) \emph{Raw feature processing (Raw):} The linear layer, which projects raw features into initial embeddings, are replaced by MoEs. 2) \emph{Encoder (Enc):} The FFN in an encoder layer is replaced by MoEs. Typically, MoEs are widely used in \emph{every-two} or \emph{last-two} layers (i.e., every or last two layers with even indices $\ell \in [0, N-1]$)~\citep{riquelme2021scaling}. 
Besides, we further attempt to use MoEs in all encoder layers. 3) \emph{Decoder (Dec):} The final linear layer of the multi-head attention is replaced by MoEs in the decoder. 
We show the average performance over 10 unseen VRPs in Fig.~\ref{moe_setting}(a). The results reveal that applying MoEs at the shallow layer (e.g., Raw) may worsen the model performance, while using MoEs in all encoder layers (Enc\_All) or decoder (Dec) can benefit the zero-shot generalization. Therefore, in this paper, we employ MoEs in both encoder and decoder to pursue a strong unified model architecture to solve various VRPs.

\textbf{Number of Experts.} We increase the number of experts in each MoE layer to 8 and 16, and compare the derived MVMoE/8E/16E models to MVMoE/4E. We first train all models using the same number (50M) of instances. After that, we also train MVMoE/8E/16E with more data and computation to explore potential better results, based on the scaling laws~\citep{kaplan2020scaling}.
In specific, we provide MVMoE/8E/16E with more data by using larger batch sizes, which linearly scale up against the number of experts (i.e., MVMoE/4E/8E/16E are trained on 50M/100M/200M instances with batch sizes 128/256/512, respectively). 
The results in Fig.~\ref{moe_setting}(b) show that increasing the number of experts with more training data further unleashes the power of MVMoE, indicating the efficacy of MoEs in solving VRPs. 

\textbf{Gating Mechanism.} We investigate the effect of different gating levels and algorithms, including three levels (i.e., node-level, instance-level and problem-level) and three algorithms (i.e., input-choice, expert-choice and random gatings), with their details presented in Appendix~\ref{app:model_structure}. As shown in Fig.~\ref{moe_setting}(c), the node-level input-choice gating performs the best, while the node-level expert-choice gating performs the worst. Interestingly, we observe that the expert-choice gating in the decoder makes MVMoE hard to be optimized. It may suggest that each gating algorithm could have its most suitable position to serve MoEs. However, after an attempt to tune this configuration (i.e., by using MoEs only in the encoder), its performance is still inferior to the baseline, with an average gap of 7.190\% on unseen VRPs.


\subsection{Additional Results}
\label{exps_analyses}
We further provide experiments and discussions on more advanced gating algorithms, training efficiency, benchmark performance, and scalability. We refer readers to more empirical results (e.g., sensitivity analyses) in Appendix \ref{app:exps}.

\textbf{Advanced Gating.} Besides the input-choice and expert-choice gating algorithms evaluated above, we further consider soft MoEs~\citep{puigcerver2023sparse}, which is a recent advanced gating algorithm. Specifically, it performs an implicit soft assignment by distributing $K$ slots (i.e., convex combinations of all inputs) to each expert, rather than a hard assignment between inputs and experts as done by the conventional sparse and discrete gating networks. Since only $K$ (e.g., 1 or 2) slots are distributed to each expert, it can save much computation. We train MVMoE on $n=50$ by using node-level soft MoEs in the decoder, following training setups.
We also show the result of employing heuristic (random) hierarchical gating in the decoder.
However, their results are unsatisfactory as shown in Fig.~\ref{moe_setting}(d).

\textbf{Training Efficiency.} Fig. \ref{moe_setting}(e) shows the training time of employing each gating algorithm in the decoder, combining with their results reported in Fig. \ref{moe_setting}(d), demonstrating the efficacy of the proposed hierarchical gating in reducing the training overhead with only minor losses in performance.

\textbf{Benchmark Performance.} We further evaluate the out-of-distribution (OOD) generalization performance of all neural solvers on CVRPLIB benchmark instances. Detailed results can be found in Tables \ref{exp_benchmark} and \ref{exp_benchmark_large_scale} in Appendix \ref{app:exps}.
Surprisingly, we observe that MVMoE/4E performs poorly on large-scale instances (e.g., $n>500$). It may be caused by the generalization issue of sparse MoEs when transferring to new distributions or domains, which is still an open question in the MoE literature~\citep{fedus2022review}. In contrast, MVMoE/4E-L mostly outperforms MVMoE/4E, demonstrating more favourable potential of the hierarchical gating in promoting the OOD generalization capability.
It is worth noting that all neural solvers are only trained on the simple uniformly distributed instances with the size $n=100$. Embracing more varied problem sizes (cross-size) and attribute distributions (cross-distribution) into the multi-task training (cross-problem) may further consolidate their performance.

\textbf{Scalability.} Given that supervised learning based approaches appear to be more scalable than RL-based approaches in the current literature, we try to build upon a more scalable method, i.e., LEHD~\citep{luo2023neural}. Concretely, we train a dense model LEHD and a light sparse model with 4 experts LEHD/4E-L on CVRP. The training setups are kept the same as~\citet{luo2023neural}, except that we train all models for only 20 epochs for the training efficiency. We use the hierarchical MoE in each decoder layer of LEHD/4E-L. The results are shown in Table~\ref{exp_benchmark_large_scale}, which demonstrates the potential of MoE as a general idea that can further benefit recent scalable methods. 
Moreover, during the solution construction process, recent works~\citep{drakulic2023bq,gao2023towards} typically constrain the search space within a neighborhood of the currently selected node, which is shown to be effective in handling large-scale instances. Integrating MVMoE with these simple yet effective techniques may further improve large-scale performance.

\section{Conclusion}
\label{conclusion}
Targeting a more generic and powerful neural solver for solving VRPs, we propose a multi-task vehicle routing solver with MoEs (MVMoE), which can solve a range of VRPs concurrently, even in a zero-shot manner. 
We provide valuable insights on how to apply MoEs in neural VRP solvers, and propose an effective and efficient hierarchical gating mechanism. 
Empirically, MVMoE demonstrates strong generalization capability on zero-shot, few-shot settings, and real-world benchmark.
Despite this paper presents the first attempt towards a large VRP model, the scale of parameters is still far less than LLMs. 
We leave 
1) the development of \emph{scalable} MoE-based models in solving \emph{large-scale VRPs}, 
2) the venture of \emph{generic} representations for different problems,
3) the exploration of \emph{interpretability} of gating mechanisms~\citep{nguyen2023demystifying,nguyen2024statistical}, and
4) the investigation of \emph{scaling laws} in MoEs~\citep{krajewski2024scaling}
to the future work.
We hope our work benefit the COP community in developing large optimization (or foundation) models\footnote{\url{https://github.com/ai4co/awesome-fm4co}}.

\section*{Acknowledgements}
This research is supported by the National Research Foundation, Singapore under its AI Singapore Programme (AISG Award No: AISG3-RP-2022-031), the Singapore Ministry of Education (MOE) Academic Research Fund (AcRF) Tier 1 grant, the National Natural Science Foundation of China (Grant 62102228), and the Natural Science Foundation of Shandong Province (Grant ZR2021QF063).
We would like to thank the anonymous reviewers and (S)ACs of ICML 2024 for their constructive comments and dedicated service to the community. Jianan Zhou would like to personally express deep gratitude to his grandfather, Jinlong Hu, for his meticulous care and love during last 26 years. Eternal easy rest in sweet slumber.

\section*{Impact Statements}
This paper presents work whose goal is to advance the field of Machine Learning. There are many potential societal consequences of our work, none which we feel must be specifically highlighted here.

\bibliography{camera_ready}
\bibliographystyle{icml2024}

\newpage
\appendix
\onecolumn

\section{Setups of VRP Variants}
\label{app:vrps}
We mainly consider five constraints as presented in Section \ref{prelim}. Our base VRP variant is CVRP, from which we derive 16 VRP variants by adding additional constraints (as listed in Table~ \ref{problem}).
The node coordinates are randomly sampled from the unit square $U(0, 1)$. Below, we provide a comprehensive description of the data generation process for each constraint.

\textbf{Capacity (C).} We follow the common settings in literature \cite{kool2018attention,kwon2020pomo}. The node demand $\delta_i$ is randomly sampled from a discrete uniform distribution $\{1, 2, \dots, 9\}$. The vehicle capacity $Q$ is set to 40 and 50 for $n=50$ and $n=100$, respectively. Before feeding into the network, the node demand is further normalized by $\delta_i'=\delta_i/Q$. During the decoding process, we mask all nodes whose demands are larger than the remaining capacity of the current vehicle.

\textbf{Open Route (O).} The open route constraint does not involve extra data generation. During the decoding process, we set the open route indicator $o_t=1$ in the dynamic feature set $\mathcal{D}_t$. As mentioned in Section \ref{prelim}, the integration of the open route constraint into others is not a simple addition. 
Concretely, 1) \emph{O w. L:} Since the vehicle does not return to the depot node, during the decoding process, we only need to mask the node $v_j$ satisfying $l_t + d_{ij} > \mathbb{L}$, where $l_t$ is the length of the current route; $d_{ij}$ is the distance between the current node $v_i$ and unmasked node $v_j$; $\mathbb{L}$ is the predefined threshold of the route length. Without the open route constraint, we should mask the node $v_j$ satisfying $l_t + d_{ij} + d_{j0} > \mathbb{L}$;
2) \emph{O w. TW:} During the decoding process, we mask the node $v_j$ satisfying $t_t + d_{ij} > l_j$, where $t_t$ is the current time (i.e., the completed time of serving the current node $v_i$); $d_{ij}$ is the distance (i.e., travel time since the vehicle speed is 1) between the current node $v_i$ and unmasked node $v_j$; $l_j$ is the end time window of $v_j$. Without the open route constraint, we need to \emph{further} mask the node $v_j$ satisfying $\max(t_t + d_{ij}, e_j) + s_j + d_{j0} > l_0$, where $e_j$ is the start time window of $v_j$; $s_j$ is the service time; $l_0$ is the end time window of the depot node $v_0$.
3) \emph{O w. C or B:} Since the demand of the depot node is 0, it does not affect the constraint satisfaction during decoding, except that the vehicle does not need to return to the depot node at the end of each route.

\textbf{Backhaul (B).} Similar to CVRP, we first generate node demands by randomly sampling from a discrete uniform distribution $\{1, 2, \dots, 9\}$. Then, following \citet{anonymous2024multitask}, we randomly select 20\% of customers as backhauls. In this paper, we consider routes that involve a mixture of linehauls and backhauls, without the presence of strict precedence constraints between them. For neural VRP solvers, to ensure the feasibility of constructed solutions, the initial customer node chosen for each vehicle should be a linehaul, with an exception being that all remaining (unvisited) nodes are backhauls.

\textbf{Duration Limit (L).} Following \citet{anonymous2024multitask}, we set the duration limit (i.e., the maximum length of each vehicle route) as $\mathbb{L}=3$, which ensures the neural solver can find feasible solutions within the unit square space $U(0, 1)$. 

\textbf{Time Window (TW).} Following \citet{li2021learning} whose data generation is similar to the procedure as in Solomon~\citep{solomon1987algorithms}, we set the time window for the depot node $v_0$ as $[e_0, l_0] = [0, 3]$, and the service time at each customer node $v_i$ as $s_i = 0.2$. The depot node does not have service time.
The time window for the customer node $v_i$ is then obtained by 1) sampling the time window center $\gamma_i \sim U(e_0 + d_{0i}, l_0-d_{i0}-s_i)$, where $d_{0i} = d_{i0}$ denotes the distance or travel time between $v_0$ and $v_i$; 2) sampling the time window half-width $h_i$ uniformly at random from $[s_i/2, l_0/3] = [0.1, 1]$; 3) setting the time window $[e_i, l_i]$ as $[\max(e_0, \gamma_i - h_i), \min(l_0, \gamma_i + h_i)]$.

\begin{table}[!ht]
  \caption{16 VRP variants with five constraints.}
  \label{problem}
  \vskip 0.1in
  \begin{center}
  \begin{small}
  \renewcommand\arraystretch{1.0}
  \resizebox{0.75\textwidth}{!}{ 
  \begin{tabular}{l|ccccc}
    \toprule
    \midrule
    & Capacity (C) & Open Route (O) & Backhaul (B) & Duration Limit (L) & Time Window (TW) \\
    \midrule
    CVRP & \checkmark & & & & \\
    OVRP & \checkmark & \checkmark & & & \\
    VRPB & \checkmark & & \checkmark & & \\
    VRPL & \checkmark & & & \checkmark & \\
    VRPTW & \checkmark & & & & \checkmark \\
    OVRPTW & \checkmark & \checkmark & & & \checkmark \\
    OVRPB & \checkmark & \checkmark & \checkmark & & \\
    OVRPL & \checkmark & \checkmark & & \checkmark & \\
    VRPBL & \checkmark & & \checkmark & \checkmark & \\
    VRPBTW & \checkmark & & \checkmark & & \checkmark \\
    VRPLTW & \checkmark & & & \checkmark & \checkmark \\
    OVRPBL & \checkmark & \checkmark & \checkmark & \checkmark & \\
    OVRPBTW & \checkmark & \checkmark & \checkmark & & \checkmark \\
    OVRPLTW & \checkmark & \checkmark & & \checkmark & \checkmark \\
    VRPBLTW & \checkmark & & \checkmark & \checkmark & \checkmark \\
    OVRPBLTW & \checkmark & \checkmark & \checkmark & \checkmark & \checkmark \\
    \midrule
    \bottomrule
  \end{tabular}}
  \end{small}
  \end{center}
\end{table}

\newpage
\section{Multi-Task VRP Solver with MoEs}
\label{app:model_structure}
We now present more details of the multi-task VRP solver with MoEs (MVMoE). We consider POMO~\citep{kwon2020pomo} as the backbone model, which is one of the most prevalent autoregressive construction-based neural solvers in VRPs. 

\subsection{Encoder}
Given a VRP instance with $n$ nodes, the static features of the $i_{\rm{th}} (i \in [0, n])$ node are $\mathcal{S}_i = \{y_i, \delta_i, e_i, l_i\}$, where we use $y_i, \delta_i, e_i, l_i$ to denote the coordinate, node demand, start and end time of the time window, respectively. A linear layer is first used to project the raw features to $d$-dimensional (i.e., $d=128$) node embeddings $h_i^{0} = \text{Linear}([y_i, \delta_i, e_i, l_i])$. Then, a stack of $N=6$ encoder layers is applied to extract the refined node embeddings $h_i^{N}$. Specifically, as shown in Fig. \ref{moe}, each encoder layer consists of a multi-head attention (MHA) layer and a feed forward network (FFN).
Following the Transformer architecture~\citep{vaswani2017attention}, their outputs are further processed by a skip-connection and instance normalization (IN),
\begin{equation}
    \tilde{h}_i = \text{IN}(h_i^{\ell} + \text{MHA}(h_i^{\ell})),
\end{equation}
\begin{equation}
\label{eq:5}
    h_i^{\ell+1} = \text{IN}(\tilde{h}_i + \text{FFN}(\tilde{h}_i)).
\end{equation}
Below, we detail the MHA and FFN layers. 1) \emph{MHA:} the MHA layer in the encoder employs a multi-head self-attention mechanism (i.e., with $A=8$ heads) to compute attention weights between each two nodes. Formally, we define dimensions $d_k$ and $d_v$ (i.e., $d_k=d_v=d/A=16$), and compute query $q_i^{\ell,a} \in \mathbb{R}^{d_k}$, key $k_i^{\ell,a} \in \mathbb{R}^{d_k}$, and value $v_i^{\ell,a} \in \mathbb{R}^{d_v}$ as follows,
\begin{equation}
\label{att:0}
    q_i^{\ell,a} = W_Q^{\ell,a} h_i^{\ell}, \quad k_i^{\ell,a} = W_K^{\ell,a} h_i^{\ell}, \quad v_i^{\ell,a} = W_V^{\ell,a} h_i^{\ell},
\end{equation}
where $W_Q^{\ell,a}, W_K^{\ell,a}, W_V^{\ell,a}$ are the parameter matrices of the $a_{\rm{th}} (a \in [1, A])$ attention head in the $\ell_{\rm{th}} (\ell \in [0, N-1])$ encoder layer. Then, the attention weight $u^{\ell,a}_{ij}$ is computed using the Softmax function to represent the correlation between the $i_{\rm{th}}$ node and the $j_{\rm{th}}$ node as follows,
\begin{equation}
\label{att:1}
    u^{\ell,a}_{ij} = \text{Softmax}\left( \frac{(q_i^{\ell,a})^T k_j^{\ell,a}}{\sqrt{d_k}} \right).
\end{equation}
By weighted message passing among nodes and aggregating outputs from multiple heads, the final MHA output for the $i_{\rm{th}}$ node at the $\ell_{\rm{th}}$ encoder layer is obtained as follows,
\begin{equation}
\label{att:2}
    h_i^{\ell,a} = \sum_{j=0}^n u^{\ell,a}_{ij} v_j^{\ell,a},
\end{equation}
\begin{equation}
\label{att:3}
    \text{MHA}(h_i^{\ell}) = [h_i^{\ell,1}, h_i^{\ell,2}, \dots, h_i^{\ell,A}] W_O^{\ell},
\end{equation}
where $W_O^{\ell}$ is the learnable parameter and $[,]$ is the concatenate operator. 2) \emph{FFN:} the FFN layer computes node-wise projections using a hidden sublayer with the dimension $d_{f} = 512$ and a ReLU activation as follows,
\begin{equation}
    \text{FFN}(\tilde{h}_i) = W_{F}^{\ell,1} \cdot \text{ReLU}(W_{F}^{\ell,0} \tilde{h}_i + b_F^{\ell,0}) + b_F^{\ell,1},
\end{equation}
where $W_{F}^{\ell,0}, W_{F}^{\ell,1}, b_F^{\ell,0}, b_F^{\ell,1}$ are learnable parameters of the FFN in the $\ell_{\rm{th}}$ layer. In this paper, we replace the FFN in each encoder layer with an MoE layer, which consists of $m$ FFNs $\{\text{FFN}_1,\dots,\text{FFN}_m\}$ and a gating network $G$, such that Eq. (\ref{eq:5}) is rewritten as,
\begin{equation}
    h_i^{\ell+1} = 
    \text{IN}(\tilde{h}_i + \sum_{j=1}^m G(\tilde{h}_i)_j \text{FFN}_j(\tilde{h}_i)).
\end{equation}

\subsection{Decoder}
The decoder takes all node embeddings $\{h_i^N\}_{i=0}^n$ and a context embedding $h_c$ as inputs. At the $t_{\rm{th}}$ decoding step, the context embedding $h_c = [h_{\tau_{t-1}}^N, \mathcal{D}_t] \in \mathbb{R}^{d+4}$ includes the embedding of the last selected node $h_{\tau_{t-1}}^N$ (initialized as the depot node $h_0^N$) and dynamic features $\mathcal{D}_t=\{c_t, t_t, l_t, o_t\}$, where $c_t, t_t, l_t, o_t$ represent the remaining capacity of the vehicle, the current time, the length of the current partial route, and the presence indicator of the open route, respectively. Next, we compute an updated context embedding $h_c'$ through a MHA layer. Note that the MHA layer in the decoder does not employ the self-attention. Concretely, the context embedding $h_c$ is used for computing query, while the node embeddings $h_i^N$ are used for computing the key and value,
\begin{equation}
    q_c^{D,a} = W_Q^{D,a} h_c, \quad k_i^{D,a} = W_K^{D,a} h_i^N, \quad v_i^{D,a} = W_V^{D,a} h_i^N,
\end{equation}
where $W_Q^{D,a}, W_K^{D,a}, W_V^{D,a}$ are the parameter matrices of the $a_{\rm{th}}$ attention head in the decoder. Then, we follow Eqs. (\ref{att:1})-(\ref{att:3}) to obtain the updated context embedding $h_c'$. Note that invalid nodes (i.e., appending them to the current partial solution may violate the problem-specific feasibility constraints) are masked such that their attention weights are set to 0 in Eq. (\ref{att:1}). Finally, the probabilities $p_i$ of valid nodes to be selected is calculated by a single-head attention as follows,
\begin{equation}
    p_i = \text{Softmax}\left( C \cdot \text{tanh}(\frac{(h_c')^T h_i^N}{\sqrt{d_k}}) \right),
\end{equation}
where $C$ is typically set to 10 to clip the logits so as to benefit the policy exploration~\citep{bello2017neural,kool2018attention}. 
In this paper, we replace the final linear layer of MHA (i.e., $W_O^{\ell}$ in Eq. (\ref{att:3})) in the decoder with an MoE layer, which consists of $m$ linear layers with parameters $\{W_O^{\ell, 1}, \dots, W_O^{\ell,m}\}$ and a gating network $G$. In this case, Eq. (\ref{att:3}) is rewritten as:
\begin{equation}
        \text{MHA}(h_i^{\ell}) = \sum_{j=1}^m G([h_i^{\ell,1}, h_i^{\ell,2}, \dots, h_i^{\ell,A}])_j 
        [h_i^{\ell,1}, h_i^{\ell,2}, \dots, h_i^{\ell,A}] W_O^{\ell, j}.
\end{equation}

\subsection{Gating Algorithms}
We present more details of the popular gating algorithms~\citep{shazeer2017,zhou2022mixture} in MoEs. Without loss of generality, we take the node-level gating as an example. Suppose that we have $m$ experts in an MoE layer, and its input has a shape of $X \in \mathbb{R}^{I\times d}$, where $I$ is the total number of nodes in a batch of inputs. The gating network $G$ takes $X$ as inputs and outputs the score matrix $H=(X\cdot W_G) \in \mathbb{R}^{I \times m}$, where $W_G$ are trainable parameters of $G$.

\textbf{Input-Choice Gating.} Each node selects Top$K$ experts based on the score matrix $H$ in the input-choice gating. The Softmax function is applied along the last dimension of $H$ to obtain the probability matrix $P$, where $P[i, j]$ denotes the probability of the $j_{\rm{th}}$ expert being selected by the $i_{\rm{th}}$ node. However, this method does not inherently ensure load balancing. For example, an expert may receive significantly more nodes than others, resulting in a dominant expert while leaving others underfitting. Following \citet{shazeer2017}, we use the noisy Top$K$ gating with the importance \& load loss to enforce a similar number of nodes received by each expert. Concretely, a learnable Gaussian noise is added to $H$ before taking the Softmax,
\begin{equation}
\label{noisy_topk}
    H' = H + \text{StandardNormal}() \cdot \text{Softplus} (X \cdot W_{noise}),
\end{equation}
\begin{equation}
    P = \text{Softmax}\left( \text{Top}K (H') \right),
\end{equation}
where $W_{noise} \in \mathbb{R}^{d\times m}$ is a learnable matrix. An example of Python code of Eq. (\ref{noisy_topk}) is shown below.
\begin{python}
# An example of Python Code of Eq. (15)
H = x @ w_gate  # w_gate is the parameter of the gating network G
noise_stddev = torch.nn.functional.softplus(x @ w_noise) + 1e-2
noisy_scores = H + (torch.randn_like(H) * noise_stddev)  # noisy_scores: H'
\end{python}
For load-balancing purposes, the goal is to ensure that each expert receives a roughly equal number of nodes. However, the number of received nodes is discrete, and hence cannot be used as the loss information to update the network through backpropagation. \citet{shazeer2017} proposes to use a smooth estimator to approximate the number of nodes assigned to each expert. The above noise term helps with load balancing by enabling the gradient backpropagation through the smooth estimator. Then, an importance \& load loss is used as the auxiliary loss $\mathcal{L}_b$ to enforce load balancing as follows,
\begin{equation}
\label{eq:importance}
    Importance(X) = \sum_{x\in X} G(x),  
\end{equation}
\begin{equation}
\label{eq:load}
    Load(X)_j = \sum_{x\in X} \Phi \left( \frac{(x \cdot W_G)_j - \varphi(H'_x, k, j)}{\text{Softplus} ((x \cdot W_{noise})_j)} \right),
\end{equation}
\begin{equation}
\label{eq:aux_loss}
    \mathcal{L}_b = CV(Importance(X))^2 + CV(Load(X))^2,
\end{equation}
where $Importance(X)$ and $Load(X)$ are vectors with the shape of $\mathbb{R}^{m}$; $\Phi$ denotes the cumulative distribution function of the standard normal distribution; $\varphi(a, b, c)$ denotes the $b_{\rm{th}}$ largest component of $a$ excluding component $c$; $H'_x$ denotes the vector corresponding to the node $x$ in $H'$; $CV$ denotes the coefficient of variation. Specifically, Eq. (\ref{eq:importance}) encourages all experts to have equal importance (i.e., the batchwise sum of gate values), and Eq. (\ref{eq:load}) ensures balanced loads among experts. We set the weight of the auxiliary loss (i.e., $\alpha$ in Eq. (\ref{eq:loss})) as 0.01. We refer more details to \citet{shazeer2017}.

\textbf{Expert-Choice Gating.} In contrast to the input-choice gating, each expert independently selects Top$K$ nodes based on the score matrix $H$ in the expert-choice gating. The Softmax function is applied along the first dimension of $H$ to obtain $P$, where $P[i, j]$ denotes the probability of the $i_{\rm{th}}$ node being selected by the $j_{\rm{th}}$ expert. It explicitly guarantees load balancing and enables a flexible allocation of computation (e.g., a node can be distributed to various number of experts). In general, $K$ is set to $\frac{I\times \beta}{m}$, where $\beta$ is the capacity factor, representing on average how many experts are utilized by a node. We set $\beta=2$ in our experiments. Note that we do not limit the maximum number of experts for each node, since it needs to solve an extra linear programming problem~\citep{zhou2022mixture} and hence increases the computational complexity.

\textbf{Random Gating.} It is a simple heuristic that randomly route inputs to experts. For example, in the problem-level gating, we randomly route a batch of inputs to $K$ experts, similar to \citet{zuo2022taming}. 

\begin{figure*}[!t]
    \vskip 0.1in
    \begin{center}
    \centerline{\includegraphics[width=0.98\columnwidth]{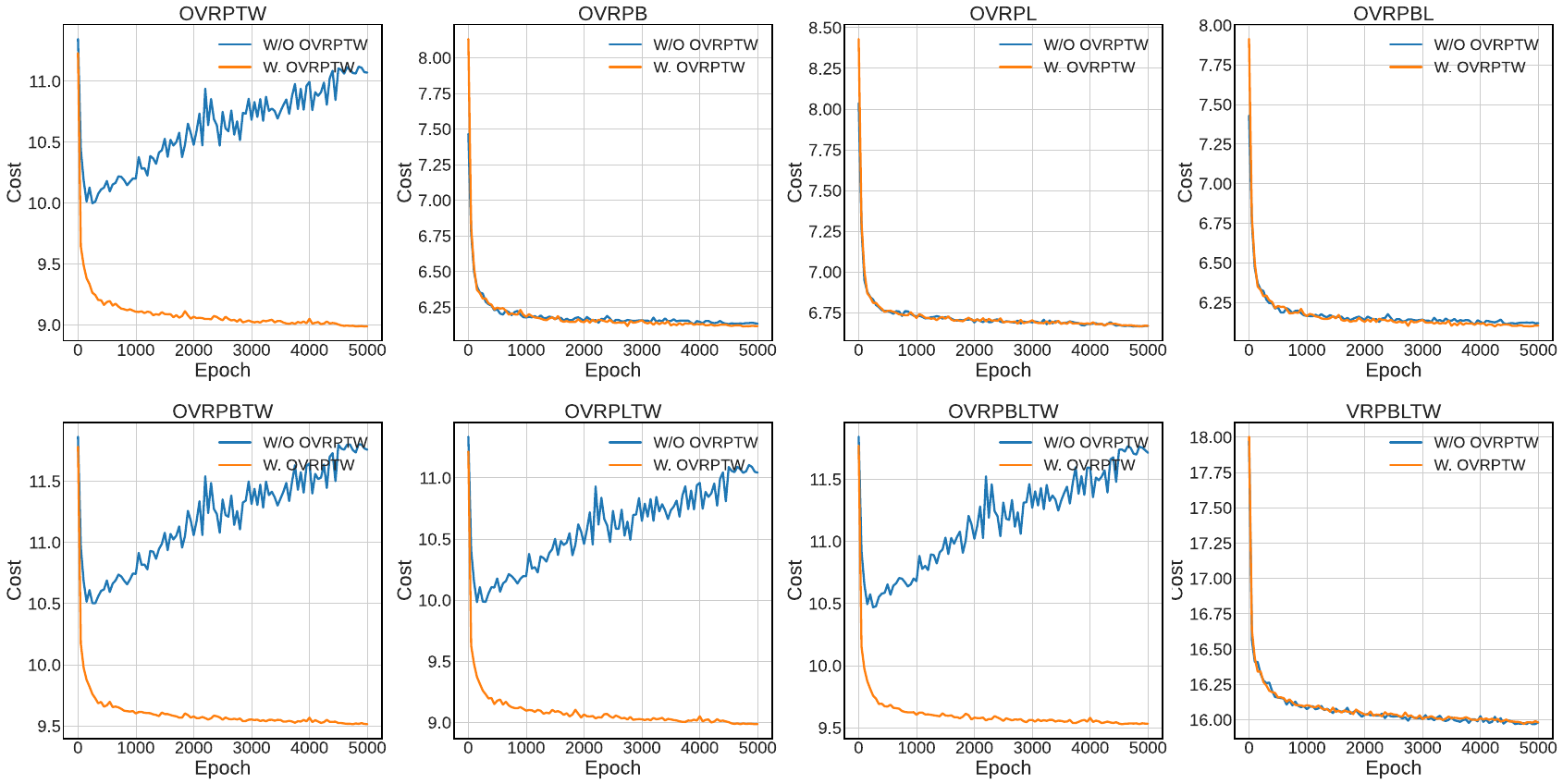}}
    \caption{The validation curves of POMO-MTL trained w/o OVRPTW and w. OVRPTW on $n=50$.}
    \label{probs}
    \end{center}
    \vskip -0.2in
\end{figure*}

\subsection{Gating Levels}
In addition to the node-level gating presented in Section \ref{methodology}, we further investigate another two gating levels in VRPs. The detailed empirical results and analyses can be found in Section \ref{exps}.

\textbf{Instance-Level Gating.} The instance-level gating routes inputs at the granularity of instances. 
Given a batch of inputs $X\in \mathbb{R}^{B\times n\times d}$, all $n$ nodes belong to the same instance are routed to the same expert(s). Before forward passing $X$ to the gating network, we apply the mean pooling operator along the second dimension of $X$, such that the score matrix has the shape of $H \in \mathbb{R}^{B\times m}$. Similar to the node-level gating, the input-choice and expert-choice gating algorithms are applicable to the instance-level gating as well. In contrast to the node-level gating, $x_i$ denotes the $i_{\rm{th}}$ instance rather than a single node in Fig. \ref{gating}. Typically, the instance-level gating can save the gating computation since $B \ll I$ as the problem scale $n$ increases. However, its empirical performance may be slightly inferior to that of the node-level gating based on our empirical results.

\textbf{Problem-Level Gating.} The problem-level gating views a batch of instances from the same VRP variant independently, and hence we route a batch of inputs $X$ to the same expert(s). Different from the node-level or instance-level gating, the input-choice and expert-choice gating algorithms are not applicable to the problem-level gating. Potential gating algorithms include randomly routing $X$ to $K$ (e.g., 1 or 2) experts~\citep{zuo2022taming} or adapting the input-choice gating without extra auxiliary losses. Although the problem-level gating is simple and efficient, it cannot guarantee all experts being sufficiently trained, since only $K$ experts are optimized in each training step. 

\section{Experiments}
\label{app:exps}
\subsection{Training Problem Set}
\label{app:train_vrps}
As mentioned in Section \ref{exps}, our training problem set includes CVRP, OVRP, VRPB, VRPL, VRPTW and OVRPTW. In contrast to our setting, the training problem set in \citet{anonymous2024multitask} does not include OVRPTW. We would like to explain this difference from two perspectives: 1) \emph{different generation of time window:} we follow \citet{li2021learning} whose data generation is similar to the procedure as in Solomon~\citep{solomon1987algorithms}, while \citet{anonymous2024multitask} follows an artificial generation process. Since Solomon dataset is based on practical constraints and real-life data, we believe its generation of time window is more realistic and challenging; 2) \emph{O meets with TW:} 
we empirically observe that the zero-shot generalization performance of POMO-MTL trained without OVRPTW becomes worse on some VRP variants, which have open route and time window constraints concurrently. Therefore, we further include OVRPTW into our training problem set to solve the challenge. The comprehensive validation curves are shown in Fig. \ref{probs}. 
We assume this empirical observation is attributed to the different data generation of time window, since it is the only difference between our settings and \citet{anonymous2024multitask}. Note that we retrain all neural methods following our training setups for a fair experimental comparison.

\subsection{Effect of Constraints}
We empirically find the time window constraint may have a bigger effect on the generalization performance. 
We show the result of POMO trained on each problem with $n=100$ in Table \ref{effect_constraints}, where we observe the average performance of POMO-VRPTW is the worst, demonstrating that the model is easier to overfit to the time window. Note that the data generation process may also affect the generalization. As shown in Appendix \ref{app:train_vrps}, modifying the data generation of time window may significantly affect the zero-shot generalization performance of the trained model on specific VRP variants. 

\begin{table*}[!ht]
  \vskip -0.1in
  \caption{The results of POMO trained on each problem with $n=100$.}
  \label{effect_constraints}
  \vskip 0.1in
  \begin{center}
  \begin{small}
  \renewcommand\arraystretch{0.9}
  \begin{tabular}{l|cccccc|c}
    \toprule
    \midrule
     & CVRP & OVRP & VRPB & VRPL & VRPTW & OVRPTW & Average \\
    \midrule
     POMO-CVRP & \textbf{1.488\%} & 20.124\% & 5.130\% & 0.105\% & 31.593\% & 42.452\% & 16.815\% \\
     POMO-OVRP & 11.698\% & \textbf{2.192\%} & 17.877\% & 10.152\% & 38.001\% & 27.734\% & 17.942\% \\
     POMO-VRPB & 2.422\% & 20.902\% & \textbf{0.995\%} & 1.050\% & 33.739\% & 42.849\% & 16.993\% \\
     POMO-VRPL & 1.490\% & 19.606\% & 5.849\% & \textbf{0.093\%} & 30.382\% & 41.164\% & 16.431\% \\
     POMO-VRPTW & 43.850\% & 80.905\% & 79.753\% & 37.869\% & \textbf{4.307\%} & 20.832\% & 44.586\% \\
     POMO-OVRPTW & 33.839\% & 58.329\% & 56.245\% & 30.257\% & 23.518\% & \textbf{2.467\%} & 34.109\% \\
    \midrule
    \bottomrule
  \end{tabular}
  \end{small}
  \end{center}
  \vskip -0.15in
\end{table*}

\subsection{Learned Gating Policy}
\begin{figure}[!ht]
    \vskip 0.05in
    \begin{center}
    \centerline{
    \includegraphics[width=0.36\columnwidth]{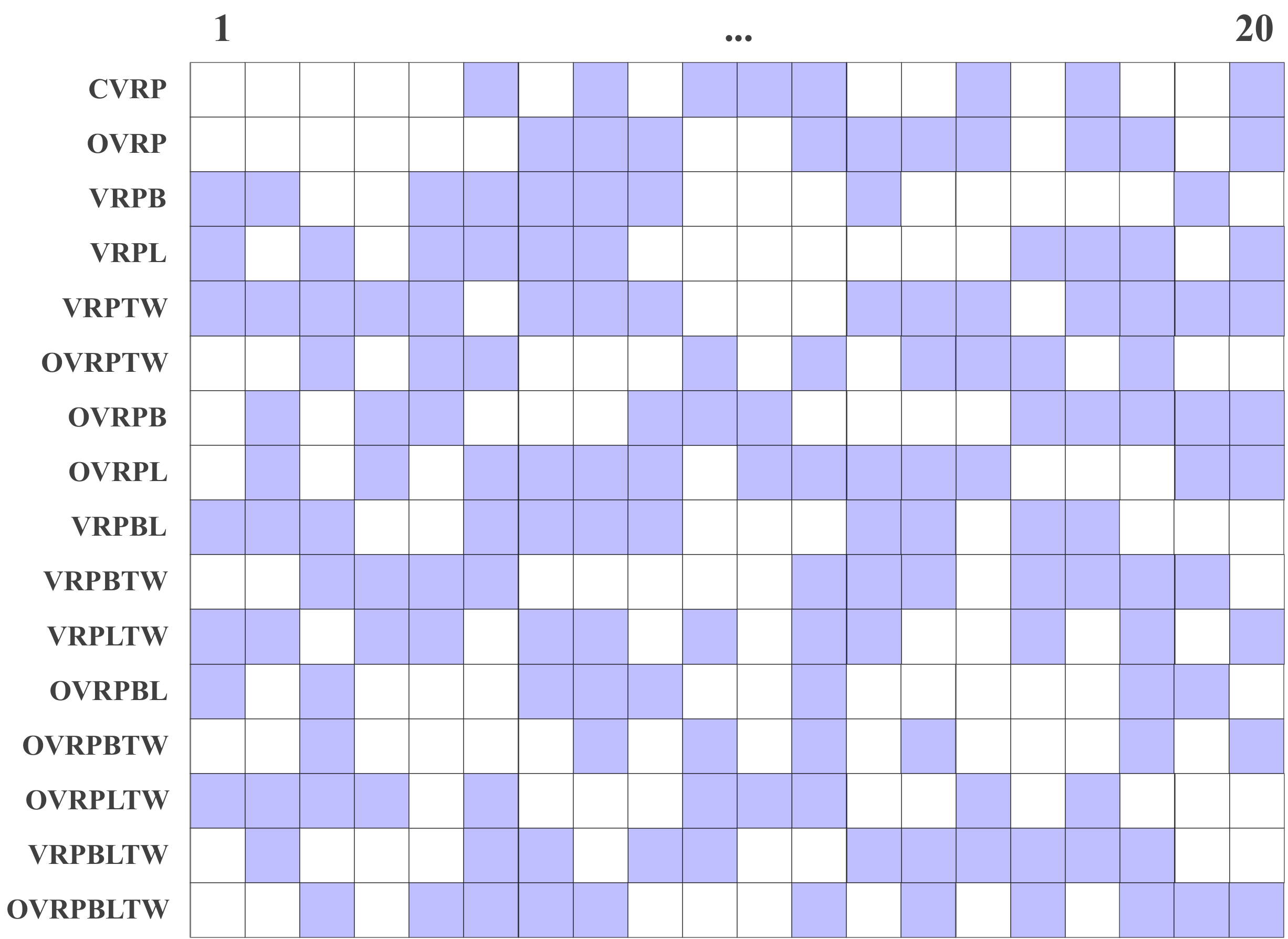}
    \hspace{3mm}
    \includegraphics[width=0.6\columnwidth]{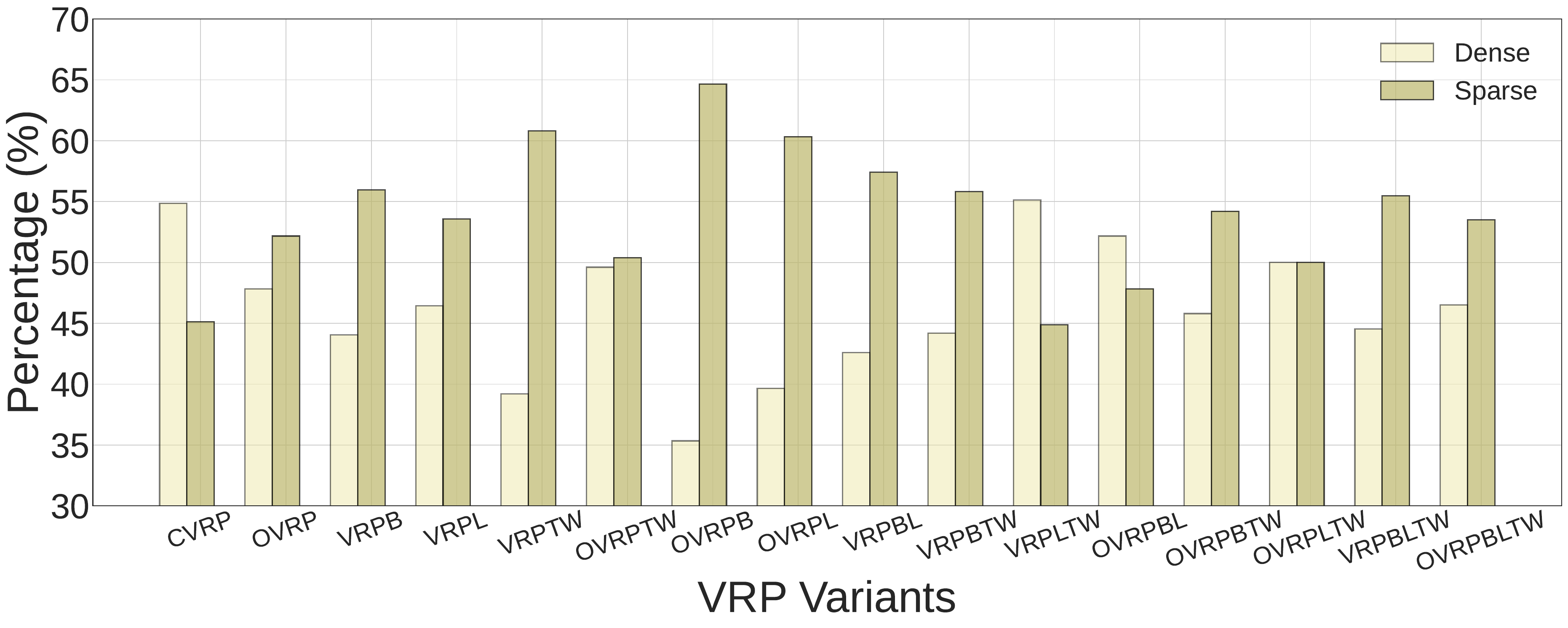} }
    \caption{\emph{Left panel:} An illustration of the first gating network's decision in MVMoE/4E-L. For simplicity, we only show the first 20 decoding steps. The colored square represents the selection of the sparse layer, while the blank square represents the selection of the dense layer.
    \emph{Right panel:} The percentage of decoding steps which route inputs to the dense or sparse layer by the first gating network.}
    \label{learned_gating}
    \end{center}
    \vskip -0.2in
\end{figure}

\textbf{Hierarchical Gating.} Here, we give a simple illustration of the learned policy of the first gating network $G_1$ in MVMoE/4E-L. In specific, in Fig. \ref{learned_gating}, we show the gating decision of $G_1$, and the percentage of decoding steps which route inputs to the dense or sparse layer, on all 16 VRP variants during inference.
Since we use the problem-level gating in the first stage, it is expected that $G_1$ explores various gating policies for different VRPs, which can be justified by the statistics in Fig. \ref{learned_gating}, in order to obtain diverse solution construction policies. 
Note that the interpretability of the learned gating policy is still a challenging task in the literature of MoEs~\citep{chen2022towards,ismail2023interpretable}. 
Since this is an early work of applying MoEs in VRPs, we concentrate on the empirical performance and leave their interpretability to the future work.

\begin{figure*}[!ht]
    \begin{center}
    \centerline{\includegraphics[width=0.9\columnwidth]{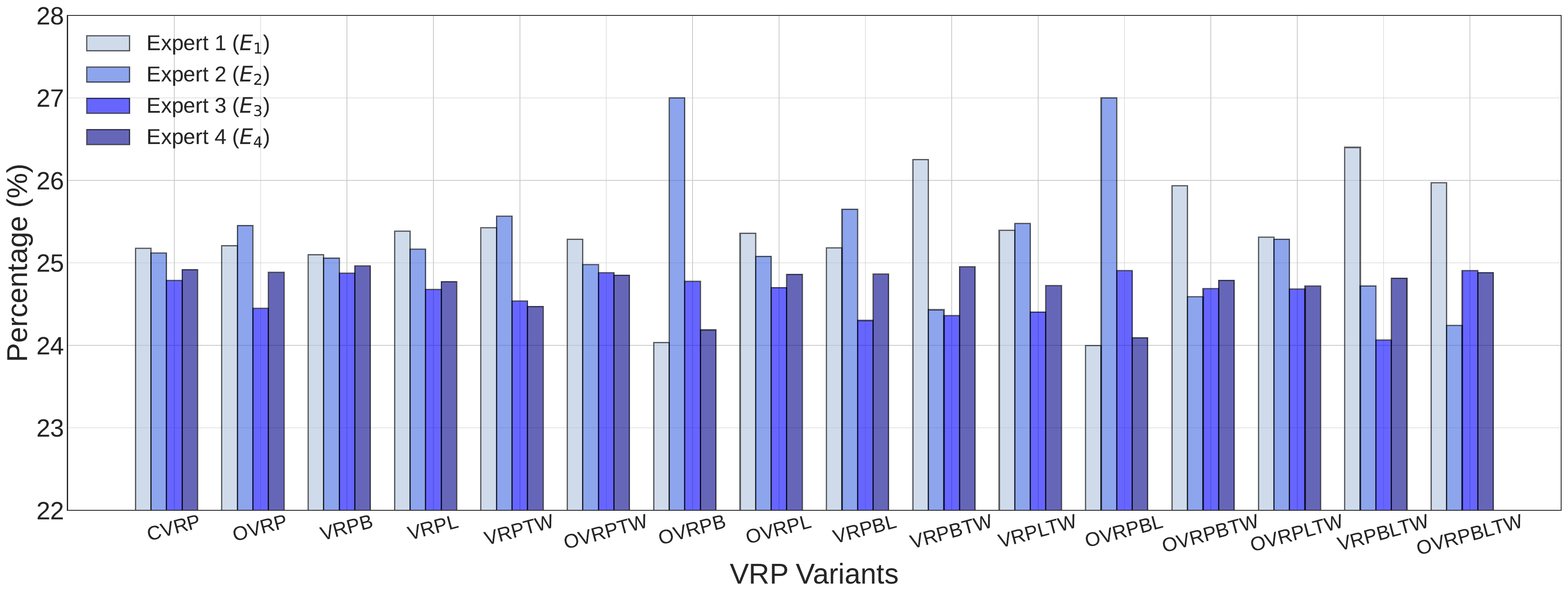}}
    \vskip -0.1in
    \caption{The average percentage of nodes being routed to each expert on 16 VRP Variants.}
    \label{loads}
    \end{center}
    \vskip -0.2in
\end{figure*}

\textbf{Expert Load.}
We expect the learned gating policy can ensure load balancing, such that each expert is sufficiently trained. Below, we take MVMoE/4E as an example, and show the load of each expert on 16 VRP variants during inference. Concretely, the load is calculated as the average percentage of nodes being routed to each expert. Based on the results in Fig. \ref {loads}, we find that 1) the loads are well-balanced among 4 experts on 6 trained VRPs; 2) the learned gating policy can generalize to unseen 10 VRPs, with only slight unbalance (e.g., maximum 27\% nodes are routed to one expert).

\subsection{Sensitivity Analyses} 
In addition to the extensive studies on MoE settings (see Section \ref{exps_moe}), we further conduct sensitivity analyses on hyperparameters, including the auxiliary loss weight $\alpha$, normalization layer, and $K$ (representing how many experts are leveraged to process each input). For the training efficiency, we follow the training setups used in Section \ref{exps_moe}, where the number of training epochs is halved on $n=50$. The detailed results can be found in Table \ref{app_sen_ana}, where we show the average performance of MVMoE/4E with different hyperparameters on the 6 trained VRPs and 10 unseen VRPs, respectively. Based on the empirical results, we find that further tuning hyperparameters (e.g., normalization layer) may boost the final performance. However, for a fair comparison with baselines, we follow their original setups~\citep{kwon2020pomo,anonymous2024multitask} by taking the first line as the default hyperparameter setting.

\begin{table}[!ht]
  \vskip -0.1in
  \caption{Sensitivity Analyses on hyperparameters.}
  \label{app_sen_ana}
  \vskip 0.1in
  \begin{center}
  \begin{small}
  \renewcommand\arraystretch{0.95}
  \begin{tabular}{ccc|cc}
    \toprule
    \midrule
    $\alpha$ & Normalization Layer & $K$ & Trained VRPs & Unseen VRPs \\
    \midrule
    0.01 & Instance Norm & 2 & 2.171\% & 6.153\% \\
    0.001 & Instance Norm & 2 & 2.156\% & 6.091\% \\
    0.1 & Instance Norm & 2 & 2.338\% & 6.438\% \\
    1.0 & Instance Norm & 2 & 2.308\% & 6.406\% \\
    0.01 & Batch Norm & 2 & 11.560\% & 18.821\% \\
    0.01 & Layer Norm & 2 & 2.048\% & 5.973\% \\
    0.01 & None & 2 & \textbf{2.033\%} & \textbf{5.842\%} \\ 
    0.01 & Instance Norm & 1 & 2.610\% & 6.751\% \\ 
    0.01 & Instance Norm & 3 & 2.126\% & 6.109\% \\ 
    \midrule
    \bottomrule
  \end{tabular}
  \end{small}
  \end{center}
  \vskip -0.1in
\end{table}

\subsection{Benchmark Performance}
We evaluate all neural solvers on CVRPLIB benchmark dataset, including CVRP and VRPTW instances with various problem sizes and attribute distributions. We mainly consider the classic Set-X~\citep{uchoa2017new} and Set-Solomon~\citep{solomon1987algorithms}.
Note that all neural solvers are only trained on the simple uniformly distributed instances with the size $n=100$ (i.e., the customer nodes' locations follow the uniform distribution). 
The comprehensive results are shown in Tables \ref{exp_benchmark} and \ref{exp_benchmark_large_scale}, where we observe 1) the single task training method (i.e., POMO) may overfit to the uniformly distributed data, and hence its out-of-distribution (OOD) generalization performance is worst; 2) the OOD generalization capability of our sparse models is generally stronger than the dense model counterpart POMO-MTL, but the superiority tends to vanish on large-scale instances; 3) surprisingly, MVMoE/4E-L performs better than MVMoE/4E, demonstrating more favourable potential of the proposed hierarchical gating in promoting the OOD generalization capability of sparse MoE-based models.

\section{Discussions}
\label{app:discussion}
\textbf{The motivation of multi-task VRP solver.} 1) From the perspective of operation research (OR): We believe generality is a favorable objective in the community. Both exact solvers (e.g., CPLEX, Gurobi) and heuristic solvers (e.g., LKH3) can solve a wide range of problems, making them a popular choice for solving VRPs (or COPs). Despite the recent trend of developing neural solvers, they still need to be tailored for each specific problem, which may consume much computation resources. For example, if training a simple model for each problem, the total training cost would be 16 models x 3 days per model, while the training cost of a unified model is only 3 days. This resource consumption cannot be neglected from a global and practical view, since we often have not much computational resources and time in practice to train many individual (and large already) models. 2) From the perspective of ML: The recent success of LLMs has demonstrated the power of foundation models. By only using text (and vision) prompts, they can solve a wide range of tasks, including code generation, text summary, or even solving optimization problems. It is also a research trend to develop a foundation model that can unify multiple tasks in CV and NLP communities. With that said, developing a general neural solver (or foundation model) in combinatorial optimization is important, and we believe it should receive attention in the NCO community, and will be a trend in future research. Moreover, given a pretrained multi-task solver, we can also adapt it to the specific downstream task through in-context learning or efficient fine-tuning, making it an attractive choice versus training from scratch.

\textbf{Why using the problem-level gating in the first stage of the hierarchical gating?} We explain this question from two perspectives: 1) \emph{generality:} we expect the proposed hierarchical gating can be applicable to any existing gating algorithms. If we use the node-level gating in the first stage, the instance-level gating algorithms will not be applicable afterwards since we cannot guarantee all nodes belong to the same instance are routed to the same (sparse or dense) layer. Similarly, if we use the instance-level gating in the first stage, the problem-level gating algorithms (e.g., THOR~\citep{zuo2022taming}) will not be applicable in the second stage;
2) \emph{efficiency:} it is possible to route a subset of inputs to the sparse layer, while distributing others to the dense layer. However, based on the Cannikin Law (i.e., Wooden Bucket Theory), the dense layer may need to wait the (relatively expensive) sparse layer before finishing the forward pass of the entire layer. The distribution of inputs at the first stage may also require extra computation if leveraging other gating levels. Therefore, for the generality and efficiency of the proposed hierarchical gating, we use the problem-level gating in the first stage.

\textbf{The effect of the hierarchical gating mechanism.} The proposed hierarchical gating mechanism is anticipated to reduce the training complexity, and significantly improve the out-of-distribution (OOD) generalization as we empirically observed. On the one hand, it is intuitive that the hierarchical gating can improve training efficiency. The solution construction (i.e., decoding) process is autoregressive in VRPs. If using the base gating, in each decoding step, the model needs to (a) handle the problem-specific feasibility constraints, (b) forward pass the inputs to the gating network, (c) distribute and forward pass each input to the selected $k$ experts. It will induce more computation overheads as the problem size scales up. Therefore, the hierarchical gating is proposed to adaptively replace the time-consuming computation of (c) with a light forward pass to a dense layer in some decoding steps. As shown in Fig. \ref{moe_setting}(e), the training time can be reduced by around 23\%. On the other hand, as shown in Tables \ref{exp_1} and \ref{exp_benchmark}, though MVMoE/4E-L (i.e., the model with the hierarchical gating) is slightly worse than MVMoE/4E (i.e., the model with the base gating) on the synthetic datasets, MVMoE/4E-L has a much stronger OOD generalization capability than MVMoE/4E on the real-world CVRPLIB dataset. We conduct further experiments and observe that the decision of the gating network in MVMoE/4E is much worse when facing OOD data. Specifically, we evaluate MVMoE/4E on uniformly distributed data with the size $n=500$, while the training instance only has a size of $n=100$. In some decoding steps, the gating network of MVMoE/4E distributes all nodes to 3 experts, while assigning few nodes to the remaining expert (e.g., the load of each expert is around [24.1\%, 49.9\%, 0.1\%, 25.9\%]). This extreme load unbalancing may be one of the main reasons for its poor OOD generalization performance. In contrast, the hierarchical gating can mitigate this issue, which may be attributed to the learned robust representation due to its regularization effect. 
Note that the generalization issue of sparse MoEs when transferring to new distributions or domains also exists in the MoE literature~\citep{fedus2022review}, and may be worthy of future exploration.

\textbf{Given the computation burden of large models, how much time does it take to solve large-scale VRPs?} We set the inference batch size as 1, and record the average time for solving a large-scale VRP instance with the size $n=1000$. The results are shown in Table~\ref{app_inference_time}, where we observe 1) large models take more time to solve a VRP instance, but they are still quite efficient compared to classical solvers; 2) the inference time of MVMoE/4E-L is generally smaller than MVMoE/4E, since MVMoE/4E-L is a light version of MVMoE/4E. Moreover, though sparse MoE models are designed to be more parameter-efficient, they still require significant computational resources, especially in terms of GPU memory (since all experts are needed to be loaded during inference). More efficient development (e.g., leveraging parallelism techniques) may be needed for future research.

\textbf{Extending zero-shot generalization to novel problems.} It is non-trivial to enable zero-shot generalization to a novel problem, whose attribute is outside the union of predefined attributes. There may be two potential ways: 1) one can design a general representation or problem formulation for VRP tasks, such as the one leverages MILP formulation~\citep{boisvert2024towards}, and then train a unified model for MILP; 2) prompt tuning~\citep{li2021prefix} or efficient fine-tuning~\citep{hu2022lora} may be potential techniques to efficiently extend the pretrained multi-task solver to a novel problem. They have already achieved superior results in other domains, and we believe this is a good research direction worthy of future exploration.

\begin{table}[!ht]
  \vskip -0.1in
  \caption{Average time to solve a large-scale VRP instance with the size $n=1000$.}
  \label{app_inference_time}
  \begin{center}
  \begin{small}
  \renewcommand\arraystretch{1.2}
  \resizebox{\textwidth}{!}{
  \begin{tabular}{l|cccccccccccccccc}
    \toprule
    \midrule
     & CVRP & OVRP & VRPB & VRPTW & VRPL & OVRPTW & OVRPB & OVRPL & VRPBL & VRPBTW & VRPLTW & OVRPBL & OVRPBTW & OVRPLTW & VRPBLTW & OVRPBLTW \\
    \midrule
    POMO-MTL & 1.28s & 1.35s & 1.35s & 1.90s & 1.45s & 1.75s & 1.46s & 1.47s & 1.50s & 1.93s & 2.20s & 1.45s & 1.74s & 1.78s & 2.01s & 1.86s \\
    MVMoE/4E & 2.14s & 2.13s & 2.21s & 2.92s & 2.37s & 3.24s & 2.28s & 2.18s & 2.34s & 2.96s & 2.89s & 2.30s & 3.08s & 2.98s & 3.10s & 3.17s \\
    MVMoE/4E-L & 2.08s & 2.06s & 2.12s & 2.83s & 2.19s & 2.87s & 2.17s & 2.17s & 2.23s & 2.92s & 2.97s & 2.26s & 2.93s & 2.96s & 3.04s & 3.02s \\
    \midrule
    \bottomrule
  \end{tabular}}
  \end{small}
  \end{center}
  \vskip -0.1in
\end{table}

\begin{table*}[!ht]
  \caption{Results on CVRPLIB instances, including Set-X~\citep{uchoa2017new} on CVRP and Set-Solomon~\citep{solomon1987algorithms} on VRPTW. All models are only trained on the uniformly distributed data with the size $n=100$. Greedy rollout is used by default.}
  \label{exp_benchmark}
  \begin{center}
  \begin{small}
  \renewcommand\arraystretch{1.5}
  \resizebox{\textwidth}{!}{ 
  \begin{tabular}{ll|cccccccc|ll|cccccccc}
    \toprule
    \midrule
    \multicolumn{2}{c|}{Set-X} & \multicolumn{2}{c}{POMO} & \multicolumn{2}{c}{POMO-MTL} & \multicolumn{2}{c}{MVMoE/4E} & \multicolumn{2}{c|}{MVMoE/4E-L} & \multicolumn{2}{c|}{Set-Solomon} & \multicolumn{2}{c}{POMO} & \multicolumn{2}{c}{POMO-MTL} & \multicolumn{2}{c}{MVMoE/4E} & \multicolumn{2}{c}{MVMoE/4E-L} \\
     Instance & Opt. & Obj. & Gap & Obj. & Gap & Obj. & Gap & Obj. & Gap & Instance & Opt. & Obj. & Gap & Obj. & Gap & Obj. & Gap & Obj. & Gap \\
    \midrule
     X-n101-k25 & 27591 & 30138 & 9.231\% & 32482 & 17.727\% & 29361 & 6.415\% & \textbf{29015} & \textbf{5.161\%} & R101 & 1637.7 & 1805.6 & 10.252\% & 1821.2 & 11.205\% & 1798.1 & 9.794\% & \textbf{1730.1} & \textbf{5.641\%} \\
     X-n106-k14 & 26362 & 39322 & 49.162\% & 27369 & 3.820\% & 27278 & 3.475\% & \textbf{27242} & \textbf{3.338\%} & R102 & 1466.6 & \textbf{1556.7} & \textbf{6.143\%} & 1596.0 & 8.823\% & 1572.0 & 7.187\% & 1574.3 & 7.345\% \\
     X-n110-k13 & 14971 & 15223 & 1.683\% & 15151 & 1.202\% & \textbf{15089} & \textbf{0.788\%} & 15196 & 1.503\% & R103 & 1208.7 & 1341.4 & 10.979\% & \textbf{1327.3} & \textbf{9.812\%} & 1328.2 & 9.887\% & 1359.4 & 12.470\% \\
     X-n115-k10 & 12747 & 16113 & 26.406\% & 14785 & 15.988\% & 13847 & 8.629\% & \textbf{13325} & \textbf{4.534\%} & R104 & 971.5 & 1118.6 & 15.142\% & 1120.7 & 15.358\% & 1124.8 & 15.780\% & \textbf{1098.8} & \textbf{13.100\%} \\
     X-n120-k6 & 13332 & 14085 & 5.648\% & 13931 & 4.493\% & 14089 & 5.678\% & \textbf{13833} & \textbf{3.758\%} & R105 & 1355.3 & 1506.4 & 11.149\% & 1514.6 & 11.754\% & 1479.4 & 9.157\% & \textbf{1456.0} & \textbf{7.433\%} \\
     X-n125-k30 & 55539 & \textbf{58513} & \textbf{5.355\%} & 60687 & 9.269\% & 58944 & 6.131\% & 58603 & 5.517\% & R106 & 1234.6 & 1365.2 & 10.578\% & 1380.5 & 11.818\% & 1362.4 & 10.352\% & \textbf{1353.5} & \textbf{9.627\%} \\
     X-n129-k18 & 28940 & \textbf{29246} & \textbf{1.057\%} & 30332 & 4.810\% & 29802 & 2.979\% & 29457 & 1.786\% & R107 & 1064.6 & 1214.2 & 14.052\% & 1209.3 & 13.592\% & \textbf{1182.1} & \textbf{11.037\%} & 1196.5 & 12.391\% \\
     X-n134-k13 & 10916 & \textbf{11302} & \textbf{3.536\%} & 11581 & 6.092\% & 11353 & 4.003\% & 11398 & 4.416\% & R108 & 932.1 & 1058.9 & 13.604\% & 1061.8 & 13.915\% & \textbf{1023.2} & \textbf{9.774\%} & 1039.1 & 11.481\% \\
     X-n139-k10 & 13590 & 14035 & 3.274\% & 13911 & 2.362\% & 13825 & 1.729\% & \textbf{13800} & \textbf{1.545\%} & R109 & 1146.9 & 1249.0 & 8.902\% & 1265.7 & 10.358\% & 1255.6 & 9.478\% & \textbf{1224.3} & \textbf{6.750\%} \\
     X-n143-k7 & 15700 & 16131 & 2.745\% & 16660 & 6.115\% & \textbf{16125} & \textbf{2.707\%} & 16147 & 2.847\% & R110 & 1068.0 & 1180.4 & 10.524\% & 1171.4 & 9.682\% & 1185.7 & 11.021\% & \textbf{1160.2} & \textbf{8.635\%} \\
     X-n148-k46 & 43448 & 49328 & 13.533\% & 50782 & 16.880\% & 46758 & 7.618\% & \textbf{45599} & \textbf{4.951\%} & R111 & 1048.7 & 1177.2 & 12.253\% & 1211.5 & 15.524\% & \textbf{1176.1} & \textbf{12.148\%} & 1197.8 & 14.220\% \\
     X-n153-k22 & 21220 & 32476 & 53.040\% & 26237 & 23.643\% & 23793 & 12.125\% & \textbf{23316} & \textbf{9.877\%} & R112 & 948.6 & 1063.1 & 12.070\% & 1057.0 & 11.427\% & 1045.2 & 10.183\% & \textbf{1044.2} & \textbf{10.082\%} \\
     X-n157-k13 & 16876 & 17660 & 4.646\% & 17510 & 3.757\% & 17650 & 4.586\% & \textbf{17410} & \textbf{3.164\%} & RC101 & 1619.8 & 2643.0 & 63.168\% & 1833.3 & 13.181\% & 1774.4 & 9.544\% & \textbf{1749.2} & \textbf{7.988\%} \\
     X-n162-k11 & 14138 & 14889 & 5.312\% & 14720 & 4.117\% & \textbf{14654} & \textbf{3.650\%} & 14662 & 3.706\% & RC102 & 1457.4 & \textbf{1534.8} & \textbf{5.311\%} & 1546.1 & 6.086\% & 1544.5 & 5.976\% & 1556.1 & 6.771\% \\
     X-n167-k10 & 20557 & 21822 & 6.154\% & 21399 & 4.096\% & 21340 & 3.809\% & \textbf{21275} & \textbf{3.493\%} & RC103 & 1258.0 & 1407.5 & 11.884\% & \textbf{1396.2} & \textbf{10.986\%} & 1402.5 & 11.486\% & 1415.3 & 12.502\% \\
     X-n172-k51 & 45607 & 49556 & 8.659\% & 56385 & 23.632\% & 51292 & 12.465\% & \textbf{49073} & \textbf{7.600\%} & RC104 & 1132.3 & \textbf{1261.8} & \textbf{11.437\%} & 1271.7 & 12.311\% & 1265.4 & 11.755\% & 1264.2 & 11.649\% \\
     X-n176-k26 & 47812 & 54197 & 13.354\% & 57637 & 20.549\% & 55520 & 16.121\% & \textbf{52727} & \textbf{10.280\%} & RC105 & 1513.7 & \textbf{1612.9} & \textbf{6.553\%} & 1644.9 & 8.668\% & 1635.5 & 8.047\% & 1619.4 & 6.980\% \\
     X-n181-k23 & 25569 & 37311 & 45.923\% & \textbf{26219} & \textbf{2.542\%} & 26258 & 2.695\% & 26241 & 2.628\% & RC106 & 1372.7 & 1539.3 & 12.137\% & 1552.8 & 13.120\% & \textbf{1505.0} & \textbf{9.638\%} & 1509.5 & 9.968\% \\
     X-n186-k15 & 24145 & 25222 & 4.461\% & 25000 & 3.541\% & 25182 & 4.295\% & \textbf{24836} & \textbf{2.862\%} & RC107 & 1207.8 & 1347.7 & 11.583\% & 1384.8 & 14.655\% & 1351.6 & 11.906\% & \textbf{1324.1} & \textbf{9.625\%} \\
     X-n190-k8 & 16980 & 18315 & 7.862\% & \textbf{18113} & \textbf{6.673\%} & 18327 & 7.933\% & \textbf{18113} & \textbf{6.673\%} & RC108 & 1114.2 & 1305.5 & 17.169\% & 1274.4 & 14.378\% & 1254.2 & 12.565\% & \textbf{1247.2} & \textbf{11.939\%} \\
     X-n195-k51 & 44225 & 49158 & 11.154\% & 54090 & 22.306\% & 49984 & 13.022\% & \textbf{48185} & \textbf{8.954\%} & RC201 & 1261.8 & 2045.6 & 62.118\% & 1761.1 & 39.570\% & 1577.3 & 25.004\% & \textbf{1517.8} & \textbf{20.285\%} \\
     X-n200-k36 & 58578 & 64618 & 10.311\% & 61654 & 5.251\% & 61530 & 5.039\% & \textbf{61483} & \textbf{4.959\%} & RC202 & 1092.3 & 1805.1 & 65.257\% & 1486.2 & 36.062\% & 1616.5 & 47.990\% & \textbf{1480.3} & \textbf{35.520\%} \\
     X-n209-k16 & 30656 & 32212 & 5.076\% & \textbf{32011} & \textbf{4.420\%} & 32033 & 4.492\% & 32055 & 4.564\% & RC203 & 923.7 & 1470.4 & 59.186\% & \textbf{1360.4} & \textbf{47.277\%} & 1473.5 & 59.521\% & 1479.6 & 60.182\% \\
     X-n219-k73 & 117595 & 133545 & 13.564\% & \textbf{119887} & \textbf{1.949\%} & 121046 & 2.935\% & 120421 & 2.403\% & RC204 & 783.5 & 1323.9 & 68.973\% & 1331.7 & 69.968\% & 1286.6 & 64.212\% & \textbf{1232.8} & \textbf{57.342\%} \\
     X-n228-k23 & 25742 & 48689 & 89.142\% & 33091 & 28.549\% & 31054 & 20.636\% & \textbf{28561} & \textbf{10.951\%} & RC205 & 1154.0 & 1568.4 & 35.910\% & 1539.2 & 33.380\% & 1537.7 & 33.250\% & \textbf{1440.8} & \textbf{24.850\%} \\
     X-n237-k14 & 27042 & 29893 & 10.543\% & \textbf{28472} & \textbf{5.288\%} & 28550 & 5.577\% & 28486 & 5.340\% & RC206 & 1051.1 & 1707.5 & 62.449\% & 1472.6 & 40.101\% & 1468.9 & 39.749\% & \textbf{1394.5} & \textbf{32.671\%} \\
     X-n247-k50 & 37274 & 56167 & 50.687\% & 45065 & 20.902\% & 43673 & 17.167\% & \textbf{41800} & \textbf{12.143\%} & RC207 & 962.9 & 1567.2 & 62.758\% & 1375.7 & 42.870\% & 1442.0 & 49.756\% & \textbf{1346.4} & \textbf{39.831\%} \\
     X-n251-k28 & 38684 & \textbf{40263} & \textbf{4.082\%} & 40614 & 4.989\% & 41022 & 6.044\% & 40822 & 5.527\% & RC208 & 776.1 & 1505.4 & 93.970\% & 1185.6 & 52.764\% & \textbf{1107.4} & \textbf{42.688\%} & 1167.5 & 50.437\% \\
    \midrule
     \multicolumn{2}{c|}{Avg. Gap} & \multicolumn{2}{c}{16.629\%} & \multicolumn{2}{c}{9.820\%} & \multicolumn{2}{c}{6.884\%} & \multicolumn{2}{c|}{\textbf{5.160\%}} & \multicolumn{2}{c|}{Avg. Gap} & \multicolumn{2}{c}{28.054\%} & \multicolumn{2}{c}{21.380\%} & \multicolumn{2}{c}{20.317\%} & \multicolumn{2}{c}{\textbf{18.490\%}} \\
    \midrule
    \bottomrule
  \end{tabular}}
  \end{small}
  \end{center}
\end{table*}

\begin{table*}[!ht]
  \caption{Results on large-scale CVRPLIB instances. All models are only trained on the uniformly distributed data with the size $n=100$. Greedy rollout is used by default, except for the last two columns (w. 100 Random Re-Construct (RRC) iterations).}
  \label{exp_benchmark_large_scale}
  \begin{center}
  \begin{small}
  \renewcommand\arraystretch{1.5}
  \resizebox{\textwidth}{!}{ 
  \begin{tabular}{ll|cccccccc|cccccccc}
    \toprule
    \midrule
    \multicolumn{2}{c|}{Set-X} & \multicolumn{2}{c}{POMO} & \multicolumn{2}{c}{POMO-MTL} & \multicolumn{2}{c}{MVMoE/4E} & \multicolumn{2}{c|}{MVMoE/4E-L} & \multicolumn{2}{c}{LEHD} & \multicolumn{2}{c}{LEHD/4E-L} & \multicolumn{2}{c}{LEHD RRC100} & \multicolumn{2}{c}{LEHD/4E-L RRC100} \\
     Instance & Opt. & Obj. & Gap & Obj. & Gap & Obj. & Gap & Obj. & Gap & Obj. & Gap & Obj. & Gap & Obj. & Gap & Obj. & Gap \\
    \midrule
     X-n502-k39 & 69226 & 75617 & 9.232\% & 77284 & 11.640\% & 73533 & 6.222\% & 74429 & 7.516\% & 71438 & 3.195\% & 71984 & 3.984\% & 70678 & 2.097\% & 70304 & 1.557\% \\
     X-n513-k21 & 24201 & 30518 & 26.102\% & 28510 & 17.805\% & 32102 & 32.647\% & 31231 & 29.048\% & 25624 & 5.880\% & 25810 & 6.648\% & 24808 & 2.508\% & 25026 & 3.409\% \\
     X-n524-k153 & 154593 & 201877 & 30.586\% & 192249 & 24.358\% & 186540 & 20.665\% & 182392 & 17.982\% & 280556 & 81.480\% & 198498 & 28.400\% & 194569 & 25.859\% & 188040 & 21.636\% \\
     X-n536-k96 & 94846 & 106073 & 11.837\% & 106514 & 12.302\% & 109581 & 15.536\% & 108543 & 14.441\% & 103785 & 9.425\% & 104750 & 10.442\% & 102098 & 7.646\% & 102414 & 7.979\% \\
     X-n548-k50 & 86700 & 103093 & 18.908\% & 94562 & 9.068\% & 95894 & 10.604\% & 95917 & 10.631\% & 90644 & 4.549\% & 88779 & 2.398\% & 88161 & 1.685\% & 88383 & 1.941\% \\
     X-n561-k42 & 42717 & 49370 & 15.575\% & 47846 & 12.007\% & 56008 & 31.114\% & 51810 & 21.287\% & 44728 & 4.708\% & 44509 & 4.195\% & 43890 & 2.746\% & 43847 & 2.645\% \\
     X-n573-k30 & 50673 & 83545 & 64.871\% & 60913 & 20.208\% & 59473 & 17.366\% & 57042 & 12.569\% & 53482 & 5.543\% & 54981 & 8.502\% & 53222 & 5.030\% & 53309 & 5.202\% \\
     X-n586-k159 & 190316 & 229887 & 20.792\% & 208893 & 9.761\% & 215668 & 13.321\% & 214577 & 12.748\% & 232867 & 22.358\% & 217229 & 14.141\% & 211415 & 11.086\% & 209941 & 10.312\% \\
     X-n599-k92 & 108451 & 150572 & 38.839\% & 120333 & 10.956\% & 128949 & 18.901\% & 125279 & 15.517\% & 115377 & 6.386\% & 117192 & 8.060\% & 112742 & 3.957\% & 113914 & 5.037\% \\
     X-n613-k62 & 59535 & 68451 & 14.976\% & 67984 & 14.192\% & 82586 & 38.718\% & 74945 & 25.884\% & 62484 & 4.953\% & 62548 & 5.061\% & 61843 & 3.877\% & 62191 & 4.461\% \\
     X-n627-k43 & 62164 & 84434 & 35.825\% & 73060 & 17.528\% & 70987 & 14.193\% & 70905 & 14.061\% & 67568 & 8.693\% & 66779 & 7.424\% & 65509 & 5.381\% & 65117 & 4.750\% \\
     X-n641-k35 & 63682 & 75573 & 18.672\% & 72643 & 14.071\% & 75329 & 18.289\% & 72655 & 14.090\% & 68249 & 7.172\% & 67355 & 5.768\% & 66053 & 3.723\% & 66196 & 3.948\% \\
     X-n655-k131 & 106780 & 127211 & 19.134\% & 116988 & 9.560\% & 117678 & 10.206\% & 118475 & 10.952\% & 117532 & 10.069\% & 113165 & 5.980\% & 112228 & 5.102\% & 110707 & 3.678\% \\
     X-n670-k130 & 146332 & 208079 & 42.197\% & 190118 & 29.922\% & 197695 & 35.100\% & 183447 & 25.364\% & 220927 & 50.977\% & 183681 & 25.523\% & 184934 & 26.380\% & 179615 & 22.745\% \\
     X-n685-k75 & 68205 & 79482 & 16.534\% & 80892 & 18.601\% & 97388 & 42.787\% & 89441 & 31.136\% & 72946 & 6.951\% & 73783 & 8.178\% & 71635 & 5.029\% & 71723 & 5.158\% \\
     X-n701-k44 & 81923 & 97843 & 19.433\% & 92075 & 12.392\% & 98469 & 20.197\% & 94924 & 15.870\% & 86327 & 5.376\% & 85860 & 4.806\% & 84330 & 2.938\% & 83965 & 2.493\% \\
     X-n716-k35 & 43373 & 51381 & 18.463\% & 52709 & 21.525\% & 56773 & 30.895\% & 52305 & 20.593\% & 46502 & 7.214\% & 47517 & 9.554\% & 45655 & 5.261\% & 46028 & 6.121\% \\
     X-n733-k159 & 136187 & 159098 & 16.823\% & 161961 & 18.925\% & 178322 & 30.939\% & 167477 & 22.976\% & 149115 & 9.493\% & 150814 & 10.740\% & 144934 & 6.423\% & 146338 & 7.454\% \\
     X-n749-k98 & 77269 & 87786 & 13.611\% & 90582 & 17.229\% & 100438 & 29.985\% & 94497 & 22.296\% & 83439 & 7.985\% & 83951 & 8.648\% & 81801 & 5.865\% & 82346 & 6.571\% \\
     X-n766-k71 & 114417 & 135464 & 18.395\% & 144041 & 25.891\% & 152352 & 33.155\% & 136255 & 19.086\% & 131487 & 14.919\% & 130899 & 14.405\% & 126293 & 10.380\% & 130511 & 14.066\% \\
     X-n783-k48 & 72386 & 90289 & 24.733\% & 83169 & 14.897\% & 100383 & 38.677\% & 92960 & 28.423\% & 76766 & 6.051\% & 77698 & 7.338\% & 75611 & 4.455\% & 76014 & 5.012\% \\
     X-n801-k40 & 73305 & 124278 & 69.536\% & 85077 & 16.059\% & 91560 & 24.903\% & 87662 & 19.585\% & 77546 & 5.785\% & 78187 & 6.660\% & 75453 & 2.930\% & 75318 & 2.746\% \\
     X-n819-k171 & 158121 & 193451 & 22.344\% & 177157 & 12.039\% & 183599 & 16.113\% & 185832 & 17.525\% & 178558 & 12.925\% & 181075 & 14.517\% & 171025 & 8.161\% & 172779 & 9.270\% \\
     X-n837-k142 & 193737 & 237884 & 22.787\% & 214207 & 10.566\% & 229526 & 18.473\% & 221286 & 14.220\% & 207709 & 7.212\% & 208760 & 7.754\% & 203657 & 5.120\% & 203254 & 4.912\% \\
     X-n856-k95 & 88965 & 152528 & 71.447\% & 101774 & 14.398\% & 99129 & 11.425\% & 106816 & 20.065\% & 92936 & 4.464\% & 93542 & 5.145\% & 91221 & 2.536\% & 90878 & 2.150\% \\
     X-n876-k59 & 99299 & 119764 & 20.609\% & 116617 & 17.440\% & 119619 & 20.463\% & 114333 & 15.140\% & 104183 & 4.918\% & 106866 & 7.620\% & 103465 & 4.195\% & 104399 & 5.136\% \\
     X-n895-k37 & 53860 & 70245 & 30.421\% & 65587 & 21.773\% & 79018 & 46.710\% & 64310 & 19.402\% & 58028 & 7.739\% & 58157 & 7.978\% & 56481 & 4.866\% & 56749 & 5.364\% \\
     X-n916-k207 & 329179 & 399372 & 21.324\% & 361719 & 9.885\% & 383681 & 16.557\% & 374016 & 13.621\% & 385208 & 17.021\% & 363744 & 10.500\% & 355168 & 7.895\% & 355635 & 8.037\% \\
     X-n936-k151 & 132715 & 237625 & 79.049\% & 186262 & 40.347\% & 220926 & 66.466\% & 190407 & 43.471\% & 196547 & 48.097\% & 172568 & 30.029\% & 169696 & 27.865\% & 163087 & 22.885\% \\
     X-n957-k87 & 85465 & 130850 & 53.104\% & 98198 & 14.898\% & 113882 & 33.250\% & 105629 & 23.593\% & 90295 & 5.651\% & 89334 & 4.527\% & 88187 & 3.185\% & 88384 & 3.415\% \\
     X-n979-k58 & 118976 & 147687 & 24.132\% & 138092 & 16.067\% & 146347 & 23.005\% & 139682 & 17.404\% & 127972 & 7.561\% & 130662 & 9.822\% & 125137 & 5.178\% & 126113 & 5.999\% \\
     X-n1001-k43 & 72355 & 100399 & 38.759\% & 87660 & 21.153\% & 114448 & 58.176\% & 94734 & 30.929\% & 76689 & 5.990\% & 75933 & 4.945\% & 75742 & 4.681\% & 75403 & 4.213\% \\
    \midrule
     \multicolumn{2}{c|}{Avg. Gap} & \multicolumn{2}{c}{29.658\%} & \multicolumn{2}{c}{\textbf{16.796\%}} & \multicolumn{2}{c}{26.408\%} & \multicolumn{2}{c|}{19.607\%} & \multicolumn{2}{c}{12.836\%} & \multicolumn{2}{c}{\textbf{9.678\%}} & \multicolumn{2}{c}{7.001\%} & \multicolumn{2}{c}{\textbf{6.884\%}} \\
    \midrule
    \bottomrule
  \end{tabular}}
  \end{small}
  \end{center}
\end{table*}

\end{document}